\documentclass[]{TEAI}
\usepackage{helvet}

\usepackage{amsmath} 
\usepackage{natbib}
\usepackage{graphicx}
\usepackage{subcaption} 

\usepackage[toc,page,header]{appendix}
\usepackage[utf8]{inputenc}
\usepackage[T1]{fontenc}
\usepackage{hyperref}
\usepackage{url}
\usepackage{booktabs}
\usepackage{amsfonts}
\usepackage{nicefrac}
\usepackage{microtype}
\usepackage{wrapfig}

\usepackage{amssymb}
\usepackage{fontawesome}
\usepackage{url}

\usepackage{titletoc}

\usepackage{tikz}
\usepackage{comment}
\usepackage{tabularx}
\usepackage{booktabs}

\usepackage{minitoc}

\usepackage{booktabs}
\usepackage{array}
\usepackage{etoolbox}

\definecolor{lightblue}{RGB}{200, 230, 255}  
\definecolor{headerblue}{RGB}{150, 200, 255} 

\usepackage{pgfplots}
\usepackage[utf8]{inputenc}
\usepackage[T1]{fontenc}
\usepackage{hyperref}
\usepackage{url}
\usepackage{booktabs}
\usepackage{amsfonts}
\usepackage{nicefrac}
\usepackage{microtype}
\usepackage{xcolor}
\usepackage{graphicx}
\usepackage{float}
\usepackage{comment}
\usepackage{multirow}
\usepackage{amsmath}
\usepackage{makecell}
\usepackage{siunitx}
\usepackage{tikz}
\usepackage{pgf-pie}
\usepackage{subcaption}
\usepackage{wrapfig}
\usepackage[export]{adjustbox}

\usepackage{ragged2e}
\usepackage{tabularx}
\usepackage{array}
\usepackage{caption}
\usepackage{enumitem}
\usepackage{pifont}
\usepackage[hang,flushmargin]{footmisc}

\usepackage{tcolorbox}

\usepackage{tcolorbox}
\tcbuselibrary{breakable}
\tcbuselibrary{skins}

\usepackage{tabularx}
\usepackage{listings}

\definecolor{LeftColor}{HTML}{66CCBE}
\definecolor{RightColor}{HTML}{E27E97}
\usepackage[table]{xcolor}

\title{Seeing Touch from Motion: \\ A Unified Modality-Aware Visuo-Tactile Policy with Tactile Motion Correlation}

\author{
    Shengqi Xu\textsuperscript{1,2,3},
    Guojin Zhong\textsuperscript{1,2},  
    Yang Liu\textsuperscript{1,2}, 
    Fanjie Wang\textsuperscript{1,2},
    Hu Luo\textsuperscript{1,2}, 
    Hanyu Zhou\textsuperscript{4},
    Weiyao Zhang\textsuperscript{1,2}
    Ziyi Ye\textsuperscript{1,2},
    Zuxuan Wu\textsuperscript{1,2,3,$\dagger$},
    Yu-Gang Jiang\textsuperscript{1,2,$\dagger$} 
}

\affiliation[1]{\mbox{Institute of Trustworthy Embodied AI, Fudan University}} 
\affiliation[2]{\mbox{Shanghai Key Laboratory of Multimodal Embodied AI}}
\affiliation[3]{\mbox{NeoteAI}}
\affiliation[4]{\mbox{School of Computing, National University of Singapore}}

\abstract{
\begin{abstract}
 VVisuo-Tactile policies leveraging optical tactile sensors have shown great promise in contact-rich manipulation. These sensors achieve high spatial resolution and multi-dimensional force sensing by utilizing  an internal camera to monitor the deformation of their elastic gel surface, thereby indirectly inferring tactile cues.
Despite their advantages, extracting fine-grained contact states necessary for contact-rich manipulation remains an open challenge. Existing methods typically use either raw images or cumulative motion fields to represent tactile cues. However, both are prone to perception ambiguity. Raw tactile images mainly capture appearance changes, while cumulative motion fields only reflect the aggregate gel deformation. Consequently, distinct fine-grained contact states can exhibit highly similar patterns, making it difficult to explicitly distinguish subtle contact variations.
To address this issue, we explore the dynamic priors of tactile motion and discover that the correlation between transient and cumulative motion can explicitly distinguish fine-grained contact states. Based on this insight, we propose a motion-aware tactile representation to facilitate contact-rich manipulation.
Beyond tactile representation, effective fusion of tactile and visual modalities is also critical. Most existing fusion methods either directly concatenate features from each modality or train modality-specific networks separately and fuse their outputs. However, these strategies struggle to simultaneously model cross-modal interactions and preserve modality-specific characteristics. In this work, we take advantage of the Mixture-of-Transformers architecture and propose a unified modality-aware visuo-tactile policy that captures cross-modal complementarity while maintaining modality-specific properties. Extensive experiments on four challenging  contact-rich manipulation tasks demonstrate the superior performance and robustness of the proposed method.
\end{abstract}
}

\checkdata[Website]{\url{https://shengqi77.github.io/Seeing-Touch-from-Motion/}}

\begin{document}
\maketitle
\renewcommand{\thefootnote}{}
\footnotetext{$^\dagger$Corresponding authors.}
\renewcommand{\thefootnote}{\arabic{footnote}}

\vspace{-1.5em}

\section{Introduction}
\label{sec:intro}

\begin{figure*}[t]
  \centering
  \captionsetup{skip=1.3mm}
     \includegraphics[width=1\linewidth]{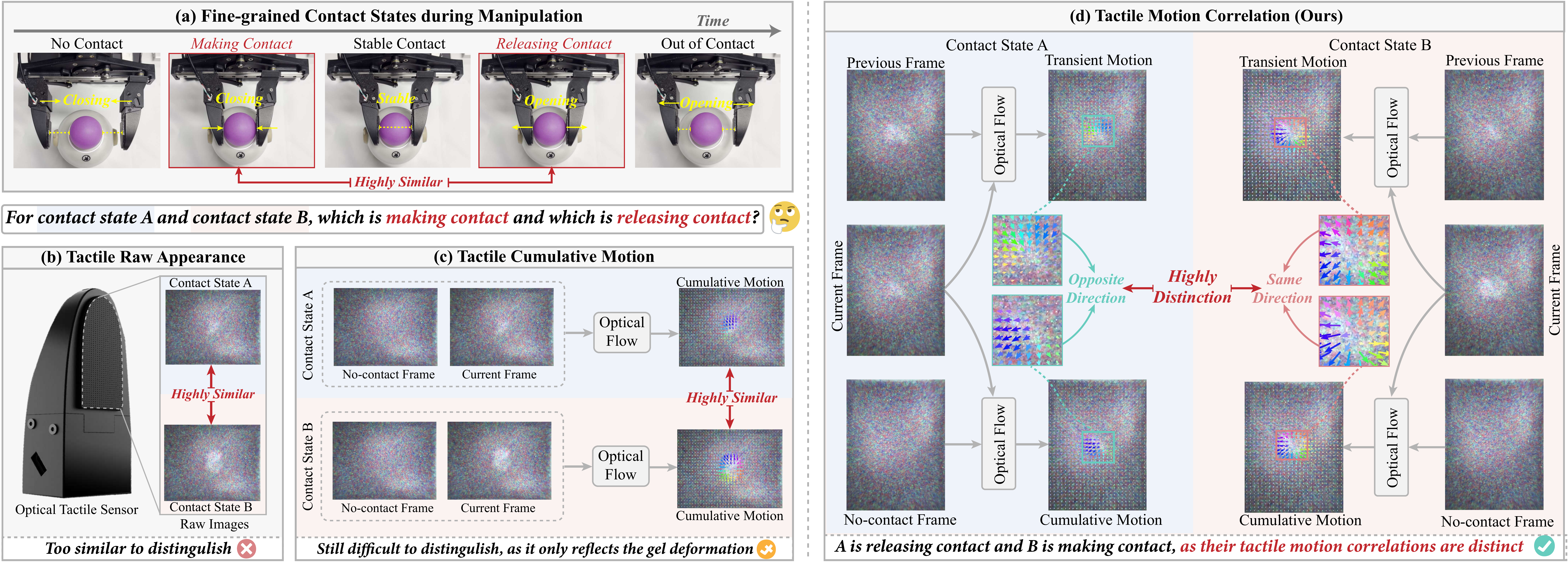}
 \caption{Comparison of three different tactile representations for distinguishing fine-grained contact states. (a)  Fine-grained contact states, such as making contact (\colorbox[HTML]{FFF4F0}{contact state B}) and releasing contact (\colorbox[HTML]{E6F0FA}{contact state A}), can be visually similar. (b) Tactile raw appearance primarily captures visual changes but suffers from ambiguity due to their deceptive similarity, making them difficult to distinguish. (c) Tactile cumulative motion (optical flow between the no-contact frame and the current frame) reflects overall deformation but still exhibits similar patterns, making it equally difficult to distinguish. (d) In contrast, our tactile motion correlation explicitly resolves this ambiguity by analyzing the dynamic relationship between transient and cumulative motion. The directional correlation (same v.s. opposite) effectively
breaks the similarity, providing highly discriminative cues crucial for contact-rich manipulation.}
  \label{Fig_1}
\end{figure*}

Contact-rich manipulation is crucial for robots performing real-world fine-grained tasks, such as precision insertion and object wiping, where success hinges on accurately perceiving subtle physical interactions. Such tasks typically rely on the complementary cooperation of global visual perception and local tactile feedback. Specifically, vision provides global contextual information to support the planning  of coarse motion trajectories, while touch captures local contact feedback to enable fine-grained interaction with objects. 

To capture local contact feedback,  a series of tactile sensing technologies has been developed.
Among these, optical tactile sensors \cite{XenseSensor,DM-TacW2,InTacS1,yuan2017gelsight,taylor2022gelslim} are widely adopted due to their high spatial resolution, texture perception and multi-dimensional force sensing capabilities. These advantages arise from their imaging mechanism, where an internal camera is utilized to monitor the deformation of an elastic gel surface. The surface is typically patterned with markers to better present gel motion during external contacts. 

Although optical tactile sensors offer information-rich visual measurements that indirectly reflect tactile cues, effectively extracting fine-grained contact states from these cues remains an underexplored problem. Most existing methods \cite{wu2025freetacman,xue2025reactive,van2024built} typically rely on either raw images or cumulative motion fields (\textit{i.e.}, optical flow between the current frame and the initial non-contact frame) to represent tactile information. However, both representations struggle to distinguish fine-grained contact states, such as the distinction between making contact and releasing contact (see Fig. \ref{Fig_1}(a)), which is crucial for contact-rich manipulation. As shown in Fig. \ref{Fig_1}(b), raw  images primarily capture appearance changes caused by extern contact but suffer from perceptual ambiguity due to their deceptive similarity, making them difficult to distinguish. Similarly, Figure \ref{Fig_1}(c) illustrates that cumulative tactile motion only reflects the direction and magnitude of overall gel deformation but still exhibits similar patterns, making it equally difficult to distinguish subtle contact variations.

To address the above issue, we thoroughly explore the underlying motion priors in tactile sensing and discover that \textit{the correlation between transient motion and cumulative motion can more explicitly distinguish fine-grained contact states}. Specifically, transient motion (\textit{i.e.}, optical flow between consecutive frames) captures instantaneous deformation, while cumulative motion captures the overall deformation relative to the initial non-contact state. We observe that different contact states exhibit distinct correlation patterns between these two types of motion. For instance,  Figure \ref{Fig_1}(d) illustrates that when releasing contact, the motion vectors point in opposite directions, whereas when making contact, they align in the same direction (See Fig. \ref{Tactile_motion}(a) for more examples). These observations inspire us to propose  \textbf{Tactile Motion Correlation (TMC)}, a motion-aware tactile representation that explicitly captures fine-grained contact states by modeling the correlation between transient and cumulative motions through their \textit{dot product}. 
This tactile representation offers three  main advantages. First, the sign and magnitude of the dot product enable precise differentiation among fine-grained contact states, while carrying a physically interpretable meaning of the underlying contact. Second, the magnitude exhibits a strong positive correlation with the applied force, thereby faithfully reflecting contact intensity. Third, it is a sensor-agnostic representation compatible with various optical sensors, allowing it to bridge  heterogeneous sensors and holds promise for enabling a general tactile representation in the future.

While an informative tactile representation is essential for contact-rich manipulation,  the effective integration of local touch with global vision is equally crucial. Existing fusion methods mainly fall into two categories: feature-level fusion  methods \cite{li2022see,zhao2024aloha,wu2025canonical} and decision-level fusion methods \cite{chen2025multi}. The former typically concatenates visual and tactile features or applies simple attention mechanisms directly to them. However, such strategies overlook the distinct properties of each modality and  may allow one modality to dominate the other. In contrast, the latter typically train an independent policy network for each modality and fuses their outputs. Though it respects modality-specific differences, the absence of interaction between vision and touch prevents the policy from exploiting cross-modal cues, making it difficult to achieve genuine complementarity.

In this work, we take advantage of the Mixture-of-Transformers (MoT) architecture \cite{liang2024mixture}  and propose \textbf{ViTacMotor}, a unified yet modality-aware \textbf{Vi}suo-\textbf{Tac}tile policy that incorporates the proposed \textbf{Mot}ion-aware c\textbf{or}relation representation. The MoT adopted in ViTacMotor decouples modality-specific parameters to preserve the unique properties of each modality, while enabling effective cross-modal interaction via a shared global self-attention mechanism. This design enables ViTacMotor to capture cross-modal complementarity while respecting the individuality of visual and tactile inputs, thereby facilitating effective integration of local touch with global vision for contact-rich manipulation. Our main contributions are summarized as follows:

\begin{enumerate}
    \item We reveal that the correlation between transient and cumulative tactile motion can more explicitly distinguish fine-grained contact states. Motivated by this finding, we propose \textbf{Tactile Motion Correlation}, a motion-aware representation that addresses the perceptual ambiguity in prior arts and thus enable fine-grained contact-state perception for facilitating contact-rich manipulation.
    \item We propose \textbf{ViTacMotor}, a unified yet modality-aware visuo-tactile policy framework. Compared to existing methods, it can capture cross-modal interactions while preserving the unique properties of each modality, enabling an effective integration of global visual perception and local tactile sensing.
    \item We compare our method with existing methods on a variety of challenging  contact-rich manipulation tasks. Extensive experiments demonstrate the superior performance and robustness of our method.
\end{enumerate}

\section{Related Work}

\textbf{Tactile Sensing in Robotics} is crucial for acquiring rich microscopic contact feedback  during interactions with objects that complements macroscopic visual perception \cite{luo2025tactile,jiang2025rotipbot,li2024vision}. Tactile sensors aim to transduce external physical contact into measurable electrical signals. They can be categorized into various types based on the transduction mechanisms, including piezoelectric \cite{zhang2022finger,huang20243d,kang2026learning}, capacitive \cite{liu2022neuro,liu2022printed}, magnetic \cite{jamone2015highly,bhirangi2021reskin}, and optical sensors \cite{yuan2017gelsight, lin20239dtact,taylor2022gelslim,lin2022dtact,ren2023mc,yamaguchi2017implementing,ward2018tactip,hogan2021seeing}. In particular, optical tactile sensors have  been widely adopted due to their ability to produce high-resolution tactile images and capture multi-dimensional force information.

Most existing visuo-tactile policies \cite{wu2025freetacman,yu2023mimictouch,li2025simultaneous,xu2025exumi,cheng2025omnivtla,bi2025vla,liu2025vitamin,george2025vital,zhang2026touchguide,zhang2025vtla,huang2026tactile}  based on optical sensors typically utilize raw tactile images as observations. For instance, Gu \textit{et al.} \cite{gu2025tactilealoha} and Wu \textit{et al.} \cite{wu2025freetacman} directly employ raw tactile images acquired from GelSight-like sensors as inputs to visuomotor policies. However, raw images mainly capture appearance changes resulting from contact and often suffer from perception ambiguity, making it challenging to reflect fine-grained contact states. Recently, several works \cite{van2024built,xue2025reactive,zhu2025residual} have attempted to employ the cumulative deformation of the gel surface as motion cues to construct tactile representation. Bogert \textit{et al.} \cite{van2024built} decompose tactile motion using Helmholtz-Hodge decomposition \cite{helmholtz1858integrale}, enabling policy transfer across different robotic embodiments. Similarly, Xue \textit{et al.} \cite{xue2025reactive} extract tactile motion cues captured by the GelSight Mini sensor and apply principal component analysis \cite{abdi2010principal} for dimensionality reduction to obtain tactile cues. However, the cumulative motion only describe the magnitude and direction of gel deformation induced by contact, making it insufficient to distinguish subtle contact variations. In this work, we reveal that the correlation of transient and cumulative motion can explicitly distinguish fine-grained contact states, and we propose tactile motion correlation as a robust tactile representation  to facilitate contact-rich manipulation. 

 A related line of learning-based tactile representation work \cite{feng2025anytouch,feng2026anytouch,higuera2024sparsh} also leverages both transient and cumulative changes, but encodes the cumulative difference and temporal tactile frames \textit{implicitly} into a latent space to  learn general tactile representation. In contrast, our method \textit{explicitly} computes the dot product between cumulative and transient motion, yielding  physically interpretable values for contact states. We appreciate their contribution and encourage readers to consult these works for complementary insights.

\noindent\textbf{Visual-Tactile Fusion in Robotics} is essential for achieving contact-rich tasks and successful interaction with the environment \cite{chen2025implicitrdp,he2025foar,huang20243d,huang2025vt,liu2025mla,zhao2025touch,yu2025forcevla}. Vision typically provides global perception, whereas tactile sensing offers local contact feedback during interactions; together they complement each other to enable closed-loop control. The most common fusion methods involve simply concatenating visual and tactile features \cite{huang2025tactile,liu2025factr,wu2025canonical,li2025adaptive} or applying straightforward attention-based fusion between the two modalities \cite{lee2025manipforce,zhu2025touch,li2022see,chen2022visuo}. Feng \textit{et al.} \cite{feng2024play} propose a stage-guided dynamic fusion that adaptively ingrates multi-modal features. Li \textit{et al.} \cite{li2022see} propose a self-attention fusion mechanism to integrate visual, auditory, and tactile modalities.  However, such strategies treat different modalities uniformly, neglecting their unique characteristics. To address this,  chen \textit{et al.} \cite{chen2025multi} introduce a policy-level approach in which separate policies are trained for each modality, and their outputs are combined via an adaptive weighting mechanism. Though this method respects the individuality of each modality, it overlooks the potential correlations between them. In this work, we propose a unified yet modality-specific visuo-tactile policy based on a Mixture-of-Transformers architecture, capturing cross-modal correlations while preserving the unique properties of each modality.

\noindent\textbf{Mixture-of-Transformers} (MoT) \cite{liang2024mixture} has recently shown strong potential in multi-modal learning. It introduces a sparse multimodal transformer architecture that leverages modality-specific parameter decoupling and global self-attention to efficiently process heterogeneous modalities while preserving cross-modal interactions. 
Owing to this capability, MoT has been adopted in various tasks, including 3D understanding and reconstruction \cite{chen2025sam}, multimodal generation and understanding \cite{wang2025hbridge,deng2025emerging,jin2025srum}, world model \cite{bi2025motus,li2026causal}, and Vision-Language-Action (VLA) model \cite{huang2025motvla,cai2026internvla,luo2026being,gu2025manualvla,wu2026pragmatic}. Closet to our method, Wu \textit{et al.} \cite{wu2026pragmatic} propose an MoT-based VLA model that performs functional decoupling: a vision–language model acts as the understanding expert, while a generative model serves as the action expert, together forming a general-purpose manipulation system. Distinctly, ViTacMotor is the first policy leveraging the MoT-based architecture to achieve unified yet modality-aware visuo-tactile fusion for contact-rich manipulation.


\section{Visuo-Tactile Policy with Tactile Motion Correlation}

\begin{figure*}[t]
  \centering
  \captionsetup{skip=1.3mm}
     \includegraphics[width=1\linewidth]{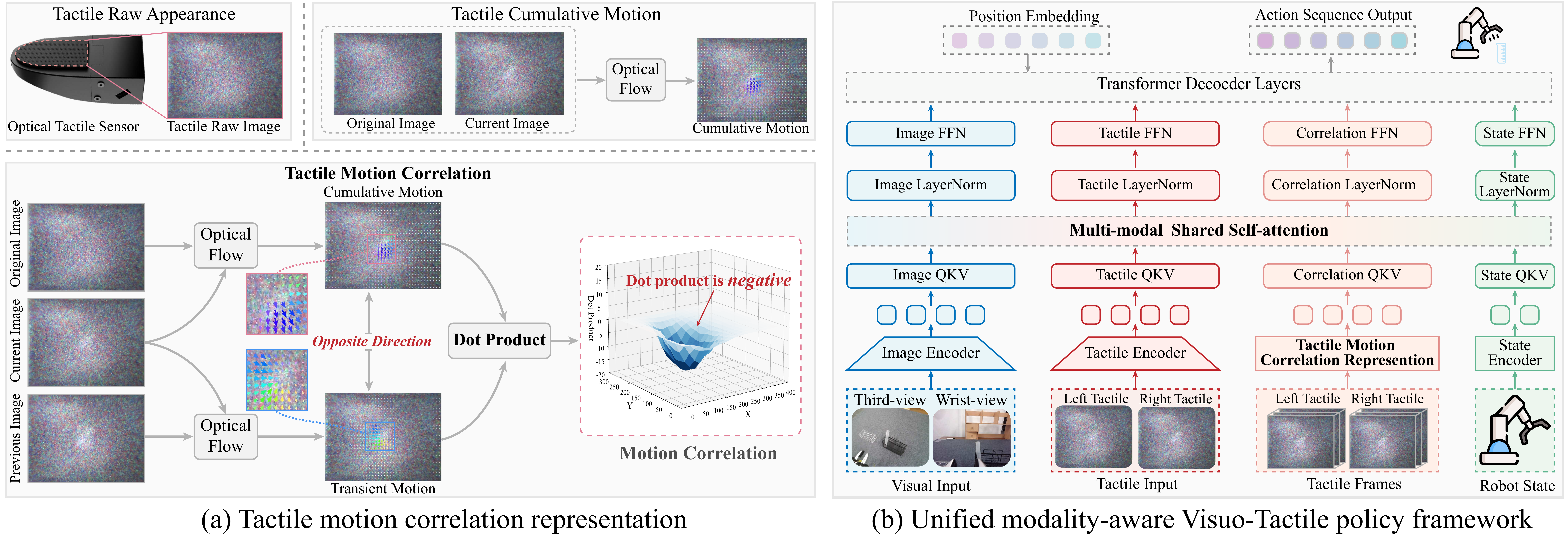}
  \caption{Overview of the Tactile Motion Correlation (TMC) representation and unified modality-aware visual-tactile policy. (a) TMC models the correlation between transient and cumulative tactile motion through their dot product, which can explicitly distinguish fine-grained contact states. (b) Unified yet modality-aware visuo-tactile policy based on MoT-based architecture for capturing cross-modal complementarity while preserving unique properties of each modality.}
  \label{Overview}
  \vspace{-2mm}
\end{figure*}

\subsection{Overview}\label{sec3.1}
In this work, we propose ViTacMotor, a unified yet modality-aware visuo-tactile policy based on a Mixture-of-Transformers (MoT) architecture, augmented with tactile motion correlation representation for contact-rich manipulation, as shown in Fig. \ref{Overview}. On the one hand, we discover that the correlation between transient and cumulative tactile motion can explicitly distinguish fine-grained contact states and propose tactile motion correlation (TMC) as a robust tactile representation to better capture fine-grained contact states (Section \ref{sec3.2}). On the other hand, we take advantage of the MoT architecture and introduce a unified yet modality-aware visuo-tactile  framework to effectively integrate global visual perception and local tactile feedback, achieving cross-modal complementarity while preserving the unique properties of each modality (Section \ref{sec3.3}). 
\subsection{Tactile Motion Correlation Representation}\label{sec3.2}
Existing visuo-tactile policies based on optical tactile sensors utilize either raw tactile appearance or cumulative motion fields as tactile information. However, both  representations struggle to distinguish fine-grained contact states, which are important for precise contact-rich manipulation.  To address this issue, we thoroughly explore the tactile motion priors of optical sensor underlying various contact states and propose a tactile motion correlation representation that more explicitly capture  such fine-grained contact states.

\noindent\textbf{Transient-Cumulative Motion Correlation.}
To explore tactile motion correlation priors underlying various fine-grained contact states (\textit{e.g.}, no contact, making contact, releasing contact, etc.), we conduct an analysis experiment using a   optical sensor  with dense markers  \cite{InTacS1}, which features an elastomer surface with markers that facilitate better capture of gel deformation. We  use an efficient optical flow estimation method \cite{kroeger2016fast} to compute two types of tactile motion: cumulative motion and transient motion. Cumulative motion refers to the optical flow between the current frame and the original non-contact frame, while transient motion is the optical flow between the consecutive frames. Note that we conduct analysis experiments on various optical sensors  \cite{DM-TacW2, XenseSensor} to validate the \textit{universality} of our findings. (See Appendix for details)

In Fig. \ref{Tactile_motion}(a), we visualize the raw tactile images, cumulative motion, and transient motion under different contact states.  Raw images mainly reflect the appearance deformation caused by contact, making it difficult to distinguish fine-grained states. Though cumulative motion captures the magnitude and direction of deformation, it provides limited discriminability, particularly between making contact, stable contact and releasing contact. In contrast, transient motion clearly captures subtle instantaneous deformations between consecutive frames. It shows clear motion differences between making contact and releasing contact, while nearly vanishing when maintaining stable contact. More importantly, we find that the correlation between transient and cumulative motion provides strong discriminative cues for contact states: (1) During no contact and out of contact, both are motionless, indicating no gel deformation; (2) During making contact, both motions exhibit the same direction, reflecting continuous accumulation of gel deformation; (3) When entering stable contact, transient motion almost disappears while cumulative motion remains, indicating the deformation has stabilized; (4) During releasing contact, the two motions show opposite directions, implying the gel is recovering. This insight reveals that \textit{transient motion complements the temporal lag of cumulative motion, and their correlation can more explicitly distinguish fine-grained contact states}, inspiring us to propose tactile motion correlation as a  tactile representation via their dot product.

\begin{figure*}[t]
  \centering
  \captionsetup{skip=1.3mm}
     \includegraphics[width=\linewidth]{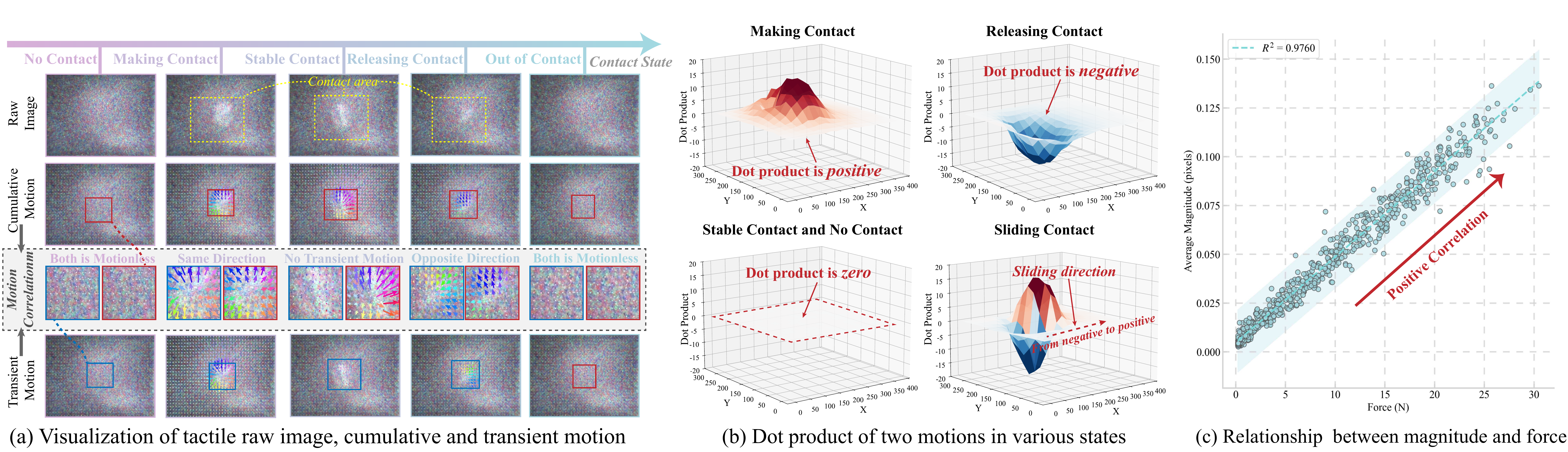}
  \caption{Analysis of tactile motion correlation properties using dense-marker optical tactile sensor \cite{InTacS1}. (a) Visualization of raw tactile images, cumulative motion, transient motion, and the correlation between the two types of motion under different fine-grained contact states. (b) 3D distribution of the dot product between transient and cumulative motion under different contact states, showing that the dot product exhibits strong discriminability across contact states. (c) Relationship between the dot product magnitude during making contact and the contact force magnitude, revealing that the dot product magnitude is positively correlated with contact force.}
  \label{Tactile_motion}
  \vspace{-2mm}
\end{figure*}

\noindent\textbf{Representing Correlation via Dot Product.} 
Given the initial frame $T_{0}$, the current frame $T_{t}$, and the previous frame $T_{t-1}$, we estimate the transient motion  $\mathbf{M}^{\mathrm{tran}}_t$ and the cumulative motion $\mathbf{M}^{\mathrm{cumu}}_t$  using optical flow:
\begin{equation}
\begin{array}{l}
\mathbf{M}^{\mathrm{tran}}_t 
= \mathbf{O}_{t}^{\,t-1 \to t}
= \mathrm{Flow}(T_{t-1}, T_t), \\[4pt]
\mathbf{M}^{\mathrm{cumu}}_t 
= \mathbf{O}_{t}^{\,0 \to t}
= \mathrm{Flow}(T_0, T_t).
\end{array}
\end{equation}
Here, $\mathrm{Flow}(\cdot, \cdot)$ denotes the optical flow estimation function. $\mathbf{O}_{t}^{\,t-1 \to t}$ denotes the instantaneous motion from the previous frame to the current frame, whereas $\mathbf{O}_{t}^{\,0 \to t}$ captures the aggregate motion relative to the initial state. To model the correlation between these two motion fields, we compute their pixel-wise dot product:
\begin{equation}
Corr_t(x)
= \mathbf{M}^{\mathrm{tran}}_t(x)
\cdot
\mathbf{M}^{\mathrm{cumu}}_t(x),
\quad x \in \Omega .
\end{equation}
where $x$ denotes a pixel location and $\Omega \subset \mathbb{Z}^2$ is the image domain. $Corr_t(x)$ refers to the motion correlation at pixel $x$ and time $t$, encoding the correlation between instantaneous and aggregate gel deformation.

\noindent\textbf{How does Dot Product Reflect Contact Feedback?} We further discuss the relationship between the dot product and the contact state. Fig. \ref{Tactile_motion}(b) illustrates the 3D distribution of the dot product between two motions under different contact states. It is observed that during making contact, the dot product is positive as the two motions are in the same direction, whereas during releasing contact it becomes negative as the two motions are opposite, and during no contact and stable contact it is close to zero as the transient motion is nearly zero. Moreover, we discover that during sliding contact, the dot product exhibits clear positive-negative separation: one side is negative while the other side is positive. The negative region corresponds to the area where the gel is recovering, while the positive region corresponds to the area where the gel is deforming, and the transition from negative to positive reflects the sliding direction. This reveals that the dot product can explicitly characterize fine-grained contact states.

Furthermore, we analyze the relationship between the dot product magnitude during making contact and the contact force in Fig. \ref{Tactile_motion}(c). We apply different levels of force to the sensor and obtain the force magnitude using the sensor SDK. It is obvious that the dot product magnitude is positively correlated with the contact force, indicating that the dot product of two motions can not only reflect fine-grained differences in contact states but also capture the magnitude of contact force, providing richer tactile cues for contact-rich manipulation.

\subsection{Unified yet Modality-aware Visuo-Tactile Fusion}\label{sec3.3}
Existing visuo-tactile fusion strategies typically follow two main paradigms: feature-level integration, which relies on direct concatenation or basic attention mechanisms, and decision-level aggregation, which involves training independent policies for each modality and subsequently averaging their outputs. However, neither strategy can simultaneously capture cross-modal complementarity and modality-specific properties. To address this, we take advantage of the Mixture-of-Transformers (MoT) architecture and propose a unified yet modality-aware fusion framework to effectively capture cross-modal complementarity and modality-specific properties. To our knowledge, we are the first to introduce the MoT architecture for visuo-tactile fusion.

\noindent\textbf{Multimodal Observation Space.} At each time step $t$, the robot perceives raw sensory inputs $\mathcal{O}_t = (I_t, T_0, T_t, T_{t-1}, p_t)$. Here, $I_t = \{ I_t^{\text{wrist}}, I_t^{\text{third}} \}$ comprises the RGB images from the wrist and third-view cameras; $T_0, T_t$, and $T_{t-1}$ represent the initial non-contact, current, and previous tactile frames, respectively; and $p_t$ denotes the robot's proprioceptive state (\textit{e.g.}, end-effector pose and gripper width).
$I_t$ and $T_t$ are processed through pre-trained encoders to extract visual embeddings $\mathbf{e}_t^{I}$ and tactile embeddings $\mathbf{e}_t^{T}$. Concurrently, we compute the transient and cumulative tactile motions using $(T_0, T_t, T_{t-1})$. The resulting tactile motion correlation representation $Corr_t$, derived as detailed in Section \ref{sec3.2}, is passed through a ResNet encoder to yield the motion correlation embedding $\mathbf{e}_t^{Corr}$. Finally, the proprioceptive state $p_t$ is projected into a latent space via an MLP encoder to obtain the proprioceptive embedding $\mathbf{e}_t^{p}$.

\noindent\textbf{Mixture-of-Transformers Multimodal Fusion.} 
Given the modality-specific embeddings $\{\mathbf{e}_t^{I}, \mathbf{e}_t^{T}, \mathbf{e}_t^{Corr}, \mathbf{e}_t^{p}\}$, we adopt an MoT-based architecture to perform a unified yet modality-aware fusion. In this framework, tactile motion correlation is treated as a distinct modality from tactile appearance to reflect their heterogeneous physical characteristics. For each modality $m \in \{I, T, Corr, p\}$, modality-specific projection matrices are applied to compute the query, key, and value vectors:
\begin{equation}
\mathbf{Q}_t^{m} = \mathbf{e}_t^{m} W_Q^{(m)}, \quad
\mathbf{K}_t^{m} = \mathbf{e}_t^{m} W_K^{(m)}, \quad
\mathbf{V}_t^{m} = \mathbf{e}_t^{m} W_V^{(m)},
\end{equation}
where $\{W_Q^{(m)}, W_K^{(m)}, W_V^{(m)}\}$ denote learnable parameters unique to each modality. Then, all modality embeddings are jointly processed through a shared self-attention mechanism to facilitate cross-modal interactions:
\begin{equation}
\mathbf{A}_t = \mathrm{softmax}\!\left( \frac{\mathbf{Q}_t\mathbf{K}_t^\top}{\sqrt{d}} \right)\mathbf{V}_t.
\end{equation}
The resulting representations are further refined through modality-specific output projections, feed-forward networks (FFN), and layer normalization (LN):
\begin{align}
\mathbf{h}_t^{m} &= \mathbf{e}_t^{m} + \mathrm{LN}_{\mathrm{attn}}^{(m)} \left( \mathbf{A}_t^{m} W_O^{(m)} \right), \\
\mathbf{y}_t^{m} &= \mathbf{h}_t^{m} + \mathrm{LN}_{\mathrm{ffn}}^{(m)} \left( \mathrm{FFN}^{(m)}(\mathbf{h}_t^{m}) \right),
\end{align}
where $W_O^{(m)}$, $\mathrm{FFN}^{(m)}$, and $\mathrm{LN}^{(m)}$ are all
modality-specific. This design allows the model to capture rich cross-modal interaction while preserving modality-specific properties.
Finally, the fused embeddings are aggregated and fed into a Transformer-based decoder to autoregressively predict the action chunk.

\vspace{-2mm}
\section{Experimental Results}

\subsection{Experimental Setup}\label{sec5.1}

\noindent\textbf{Training Details.}  Our model is trained using a $\beta$-VAE objective to model the stochasticity in human demonstrations. 
The overall loss function is defined as:
\begin{equation}
\mathcal{L} = \mathcal{L}_{\text{pred}} + \beta \mathcal{L}_{\text{KL}}.
\end{equation}
The prediction loss employs an L1 loss for precise action prediction:
\begin{equation}
\mathcal{L}_{\text{pred}} =
\left\| \hat{a}_{t:t+k} - a_{t:t+k} \right\|_1,
\end{equation}
where $\hat{a}_{t:t+k}$ denotes the predicted action chunk and $a_{t:t+k}$ denotes the ground-truth action sequence. 
The KL loss regularizes the posterior distribution
\begin{equation}
q_\phi(z \mid a_{t:t+k}, o_t)
\end{equation}
toward a standard Gaussian prior:
\begin{equation}
\mathcal{L}_{\text{KL}} =
D_{\text{KL}}\left(
q_\phi(z \mid a_{t:t+k}, o_t)
\;\|\;
\mathcal{N}(0, I)
\right).
\end{equation}
The hyperparameter $\beta$ modulates the strength of the information bottleneck.The model is trained from scratch for each task using the Adam  optimizer \cite{kingma2014adam}. 
Training takes approximately 5 hours per task on a single RTX 3090 Ti GPU. The batch size is set to 16, and the learning rate is initialized at $2\times 10^{-4}$ with a cosine decay schedule.
More implementation details are provided in the appendix.

\begin{figure*}[t]
  \centering
  \captionsetup{skip=1.3mm}
     \includegraphics[width=\linewidth]{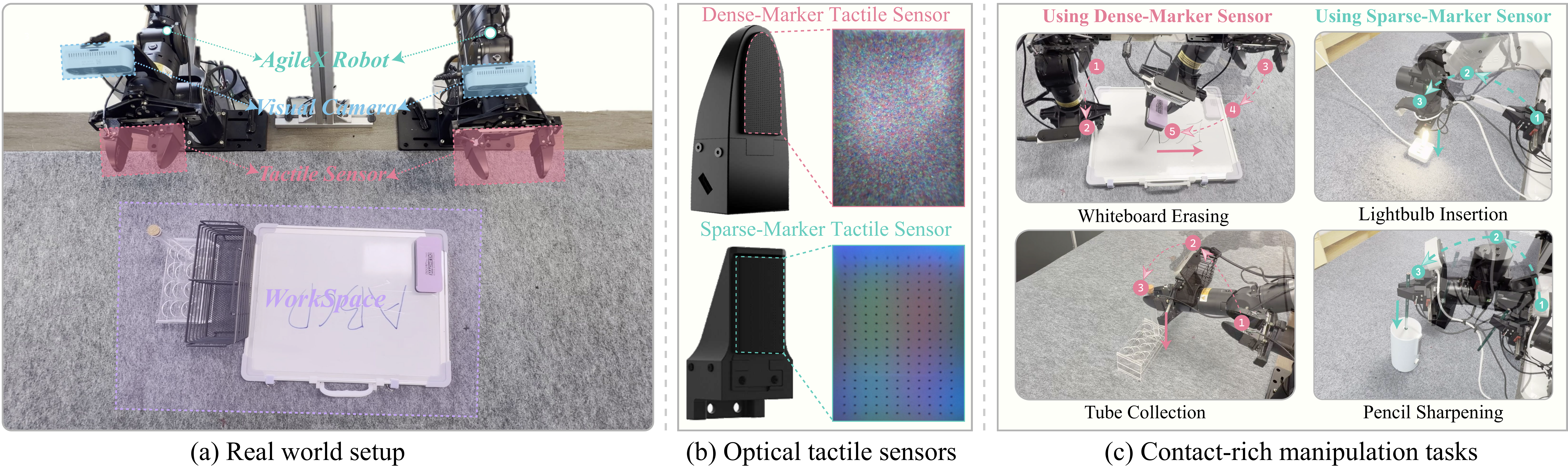}
  \caption{Illustration of the real-world experimental setup. (a) An Agilex robot  equipped with visual cameras (wrist and third-view) and  optical tactile sensors. (b)  Two kinds of representative optical tactile sensors: one with random-distributed dense markers \cite{InTacS1,DM-TacW2} and one with regularly-arranged sparse markers \cite{XenseSensor}. (c) Four challenging contact-rich manipulation tasks: cylinder collection, whiteboard erasing, lightbulb insertion, and pencil sharpening.} 
  \label{setup}
      \vspace{-2mm}
\end{figure*}

\begin{figure*}[t]
  \centering
  \captionsetup{skip=1.3mm}
     \includegraphics[width=\linewidth]{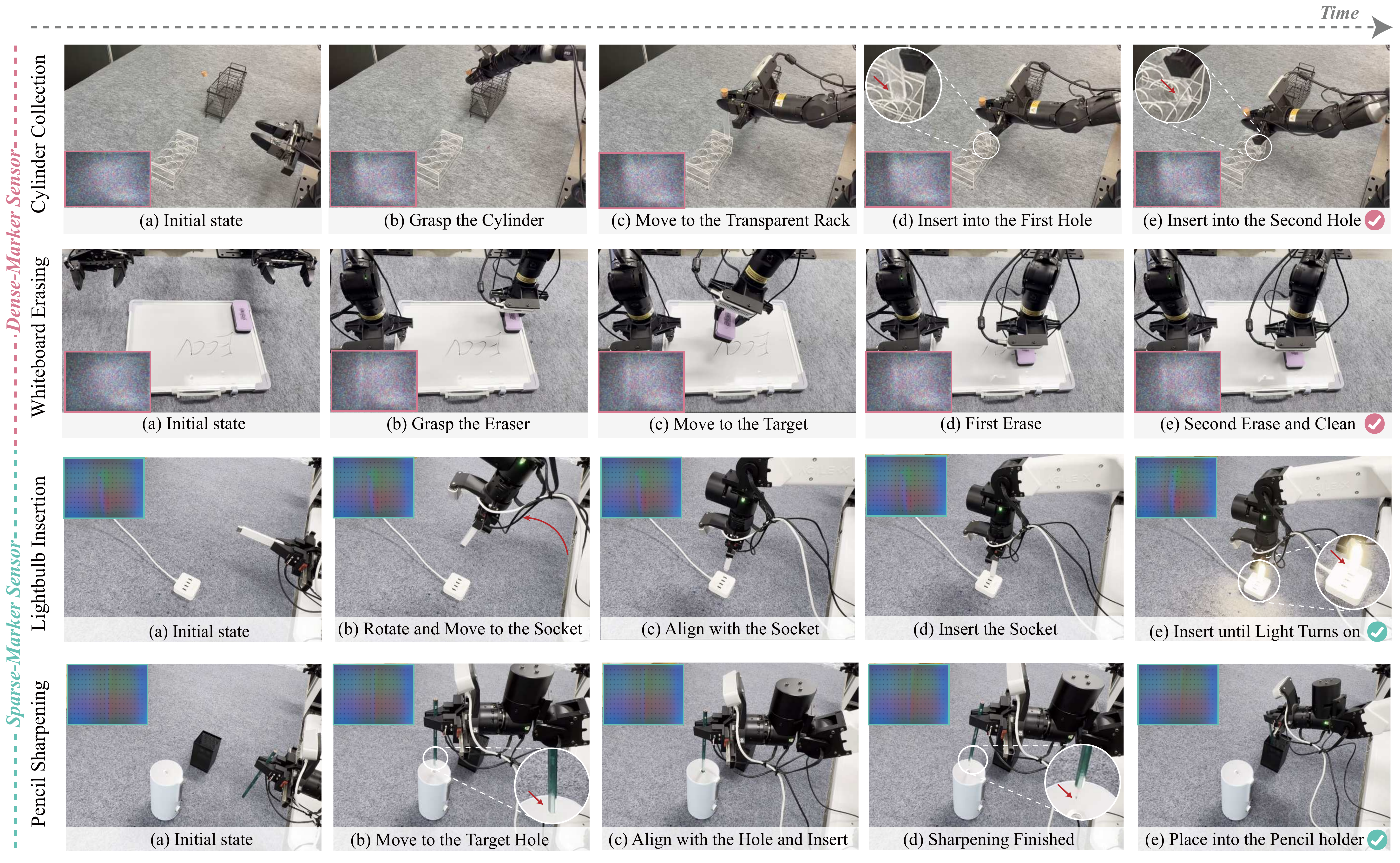}
  \caption{Qualitative results of policy execution. We evaluate our visuo-tactile policy using two kinds of optical sensors with different marker patterns on four challenging real-world contact-rich manipulation tasks: tube collection, whiteboard erasing, lightbulb insertion, and pencil sharpening.}
  \label{Fig5}
    \vspace{-2mm}
\end{figure*}

\noindent\textbf{Hardware.}  
Our robotic system  consists of two 6-DoF Agilex robotic arms, each with a 1-DoF gripper mounted on its end-effector. 
One third-view camera and two wrist-view cameras provide RGB observations for global visual perception. 
For local tactile sensing, we employ two kinds of representative tactile sensors: one with random-distributed dense markers \cite{InTacS1,DM-TacW2} and one with regularly-arranged sparse markers \cite{XenseSensor}. 
The overall system is shown in Fig.~\ref{setup}(a), and the tactile sensors are presented in Fig.~\ref{setup}(b).

\begin{table*}[t]
  \setlength{\tabcolsep}{1.5pt}
      \renewcommand\arraystretch{1.1}

\centering
\caption{Quantitative Comparison of success rates ($\%$) with Baselines. We evaluate our policy over 15 episodes, and the best performance is highlighted in bold. The numbers in parentheses denote the number of training demonstrations.}

\footnotesize
\renewcommand{\arraystretch}{1.15}
\begin{tabularx}{\textwidth}{l *{7}{>{\centering\arraybackslash}X}}
\Xhline{1pt}
\multicolumn{8}{c}{Tasks Requiring In-Hand State Information} \\
\Xhline{1pt}
\multirow{2}{*}{\textbf{Methods}} & 
\multicolumn{4}{c}{\cellcolor{RightColor!10}\textbf{Tube Collection (35 demos)}} & 
\multicolumn{3}{c}{\cellcolor{LeftColor!10}\textbf{Lightbulb Insertion (60 demos)}} \\
\cmidrule[0.8pt](lr){2-5} \cmidrule[0.8pt](lr){6-8}
& \textbf{\makecell{Grasp \\ Tube}} & \textbf{\makecell{Insert \\ 1st Hole}} & \textbf{\makecell{Insert \\ 2nd Hole}} & \textbf{\makecell{Whole \\ Task}} & \textbf{\makecell{Align with \\ Socket}} & \textbf{\makecell{Insert \& \\ Light on}} & \textbf{\makecell{Whole \\ Task}} \\
\Xhline{1pt}
ACT$^*$ & 66.7 & 60.0 & 53.3 & 53.3 & \textbf{60.0} & 13.3 & 13.3\\
DP$^*$ & 73.3 & 53.3 & 40.0 & 40.0 & 53.3 &  6.7& 6.7 \\
\Xhline{1pt}
ACT + T & 86.7 & 60.0 & 60.0 & 60.0 & \textbf{60.0} & 26.7 & 26.7 \\
Policy Consensus & 80.0 & 53.3 & 53.3 & 53.3& 53.3 & 26.7 & 26.7 \\
TactileACT & \textbf{93.3} & 73.3 & 60.0 & 60.0 & \textbf{60.0} & \textbf{40.0} & \textbf{40.0} \\
\Xhline{1pt}
\rowcolor[RGB]{239,243,252}\textbf{ViTacMotor} & \textbf{93.3} & \textbf{80.0} & \textbf{73.3} & \textbf{73.3} & \textbf{60.0} & \textbf{40.0} & \textbf{40.0} \\
\Xhline{1pt}
\multicolumn{8}{c}{Tasks Requiring Fine-Grained Force Control} \\
\Xhline{1pt}
\multirow{2}{*}{\textbf{Methods}} & 
\multicolumn{4}{c}{\cellcolor{RightColor!10}\textbf{Whiteboard Erasing (50 demos)}} & 
\multicolumn{3}{c}{\cellcolor{LeftColor!10}\textbf{Pencil Sharpening (40 demos)}} \\
\cmidrule[0.8pt](lr){2-5} \cmidrule[0.8pt](lr){6-8}
& \textbf{\makecell{Grasp \\ Eraser}}& \textbf{\makecell{First \\ Erase}} & \textbf{\makecell{Second \\ Erase}} & \textbf{\makecell{Whole \\ Task}} & \textbf{\makecell{Insert \& \\ Sharpening}} & \textbf{\makecell{Put into \\ holder}} & \textbf{\makecell{Whole \\ Task}} \\
\Xhline{1pt}
ACT$^*$ & \textbf{100.0} & 60.0 & 46.7 & 46.7 & 40.0 & 33.3 & 33.3 \\
DP$^*$ & \textbf{100.0} & 66.7 & 40.0& 40.0 & 33.3 & 33.3 & 33.3 \\
\Xhline{1pt}
ACT + T & \textbf{100.0} & 80.0 & 60.0 & 60.0 & 40.0 & 40.0 & 40.0 \\
Policy Consensus & 93.3 & 73.3 & 66.7 & 66.7 & 33.3 & 33.3 & 33.3 \\
TactileACT & \textbf{100.0} & \textbf{86.7} & 73.3 & 73.3  & 46.7 & 46.7 & 46.7 \\
\Xhline{1pt}
\rowcolor[RGB]{239,243,252} \textbf{ViTacMotor} & \textbf{100.0} & \textbf{86.7} & \textbf{86.7} & \textbf{86.7} & \textbf{60.0} & \textbf{60.0} & \textbf{60.0} \\
\Xhline{1pt}
\end{tabularx}
\label{tab:comparison}
\end{table*}

\noindent\textbf{Tasks and Evaluation Criteria.}  
We evaluate our ViTacMotor on four challenging real-world contact-rich manipulation tasks. Representative execution results for each task are shown in Fig. \ref{Fig5}. Below are the detailed descriptions and evaluation criteria for each task:

\noindent\textit{\noindent(1) Tasks Requiring Precise Contact State Information}

\noindent\textit{\textbf{Tube Collection} (Using Dense-Marker Sensor)}: The robot needs to pick up a transparent tube from a bucket and place it into a transparent rack. The task must be executed carefully to prevent breakage, with precise alignment to the correct slot and successful insertion through the two holes.
\textit{Evaluation Criteria}: A trial is considered successful if the tube is securely placed in the rack without damage.

\noindent\textit{\textbf{Lightbulb Insertion} (Using Sparse-Marker Sensor)}: The robot needs to rotate a USB-type lightbulb to the correct orientation, move it toward the socket, and insert it until the light turns on. The task requires precise alignment and adjustment during contact to ensure correct insertion. \textit{Evaluation Criteria}: A trial is deemed successful if the lightbulb is correctly inserted into the socket and illuminates.

\noindent\textit{\noindent(2) Tasks Requiring fine-grained force Control}

\noindent\textit{\textbf{Whiteboard Erasing} (Using Dense-Marker Sensor)}: The robot needs to pick up a eraser and then remove the writing from a whiteboard.  It must apply appropriate pressure to effectively erase the marker ink while avoiding excessive force that could potentially damage the system. The task requires stable and well-controlled contact force throughout. \textit{Evaluation Criteria:} The task is considered successful if all visible ink on the whiteboard is completely removed.

\noindent\textit{\textbf{Pencil Sharpening} (Using Sparse-Marker Sensor)}: The robot needs to insert a pencil into a sharpener to sharpen it, and then place it into a holder. The task requires precise alignment during contact, along with sufficient downward force to ensure effective sharpening. \textit{Evaluation Criteria}: A trial is considered successful if the pencil is successfully sharpened and then placed into the pencil holder.

\noindent\textbf{Baselines.}   
We compare our method with the following open-source baselines:

\noindent\ $\bullet$ \textit{{Action Chunking  with Transformers (ACT)}} \cite{zhao2023learning}: A transformer-based visual policy that relies only on visual images and robot states.

\noindent\ $\bullet$ \textit{ACT + Tactile Image (ACT+T)}: The ACT policy augmented with tactile images as input. The tactile images are processed by a ResNet-18 encoder and then fused with visual features using a concatenation operation.

\noindent\ $\bullet$ \textit{{Diffusion Policy (DP)}} \cite{chi2025diffusion}:  A diffusion-based visual policy that relies only on visual images and robot states. 

\noindent\ $\bullet$ \textit{Policy Consensus} \cite{chen2025multi}: A diffusion-based visuo-tactile policy that fuses vision and touch at the decision level.

\noindent\ $\bullet$ \textit{TactileACT} \cite{george2025vital}: A transformer-based visuo-tactile policy that fuses vision and touch at the feature level using a concatenate operation.

For each of the four  tasks, we conduct 15 trials for each method and report the average success rate in Table \ref{tab:comparison}.

\vspace{-3mm}
\subsection{Comparison Results}\label{sec5.2}

\noindent\textbf{Comparison on Tasks Requiring In-Hand State Information.} In Table \ref{tab:comparison}, we compare ViTacMotor with several baselines on tasks that require precise contact state information. Visual-only policies perform poorly, as visual inputs are often severely occluded. For example, in the tube collection task, when the gripper holds the tube above the rack, the third-person view is blocked by the robotic arm and gripper, making it difficult to infer the current contact state or even determine whether the tube has been inserted into the hole. This ambiguity confuses vision-based policies and degrades insertion accuracy. A similar issue arises in the lightbulb insertion task. 
Although policies augmented with raw tactile images show some improvement, they still lack robustness, as fine-grained contact states are difficult to directly interpret from raw tactile appearance. In contrast, our method explicitly perceives fine-grained contact states and achieves superior performance on these tasks.

\noindent\textbf{Comparison on Tasks Requiring Fine-Grained Force Control.}
We further compare on tasks that demand precise force control in Table \ref{tab:comparison}. Visual-only policies struggle in such scenarios, as visual observations alone cannot provide feedback on contact forces, thereby limiting precise force modulation.
For instance, in the whiteboard erasing task, the robot must apply appropriate pressure to effectively remove ink. However, vision-based methods cannot perceive contact forces and therefore fail to achieve complete erasure. In the pencil sharpening task, when the pencil is misaligned with the hole during insertion, visual inputs cannot detect the resulting shear forces, making fine corrective adjustments difficult.
Although existing visuo-tactile policies achieve comparable performance, their lack of explicit force modeling limits stable force control. In contrast, our proposed TMC method not only captures fine-grained contact states but also implicitly encodes contact force magnitudes, leading to superior performance.

\vspace{-3mm}
\subsection{Ablation and Discussion}\label{sec5.3}

\begin{table*}[t]
  \setlength{\abovecaptionskip}{4pt}
  \setlength{\belowcaptionskip}{2pt}
  \centering
  \begin{minipage}[t]{0.32\textwidth}
    \centering
    \caption{Ablation study of ViTacMotor.}
    \label{albation}
    \footnotesize
    \setlength\tabcolsep{4pt}
    \renewcommand\arraystretch{1.2}
    \begin{tabular}{ccc|cc}
      \Xhline{1pt}
      Base & TMC & \makecell{MoT-\\Fusion} & \makecell{White.\\Erasing} & \makecell{Tube\\Collect.} \\
      \hline
      \ding{51} &           &           & 46.7 & 53.3 \\
      \ding{51} & \ding{51} &           & 73.3 & 66.7 \\
      \ding{51} &           & \ding{51} & 66.7 & 60.0 \\
      \hline
      \rowcolor[RGB]{239,243,252} \ding{51} & \ding{51} & \ding{51} & \textbf{86.7} & \textbf{73.3} \\
      \Xhline{1pt}
    \end{tabular}
  \end{minipage}
  \hfill
  \begin{minipage}[t]{0.30\textwidth}
    \centering
    \caption{Effectiveness of TMC embedded into existing policies.}
    \label{embedded}
    \footnotesize
    \setlength\tabcolsep{8pt}
    \renewcommand\arraystretch{1.2}
    \begin{tabular}{l|cc}
      \Xhline{1pt}
      Methods & \makecell{ACT\\\cite{zhao2023learning}} & \makecell{DP\\\cite{chi2025diffusion}} \\
      \hline
      Raw Image    & 60.0 & 53.3 \\
      Cumu. Motion & 53.3 & 40.0 \\
      \hline
      \rowcolor[RGB]{239,243,252} \textbf{TMC (Ours)} & \textbf{73.3} & \textbf{66.7} \\
      \Xhline{1pt}
    \end{tabular}
  \end{minipage}
  \hfill
  \begin{minipage}[t]{0.35\textwidth}
    \centering
    \caption{Effectiveness of MoT-based fusion strategy compared to other methods.}
    \label{Text}
    \footnotesize
    \setlength\tabcolsep{5pt}
    \renewcommand\arraystretch{1.2}
    \begin{tabular}{l|cc|c}
      \Xhline{1pt}
      \makecell{Fusion\\Strategy} & \makecell{White.\\Erasing} & \makecell{Tube\\Collect.} & Avg. \\
      \hline
      Concatenation   & 73.3 & 66.7 & 70.0 \\
      Cross Attention & 73.3 & 60.0 & 66.7 \\
      \hline
      \rowcolor[RGB]{239,243,252} \textbf{Ours} & \textbf{86.7} & \textbf{73.3} & \textbf{80.0} \\
      \Xhline{1pt}
    \end{tabular}
  \end{minipage}
\end{table*}

\begin{figure*}[t]
  \centering
  \captionsetup{skip=1.3mm}
     \includegraphics[width=\linewidth]{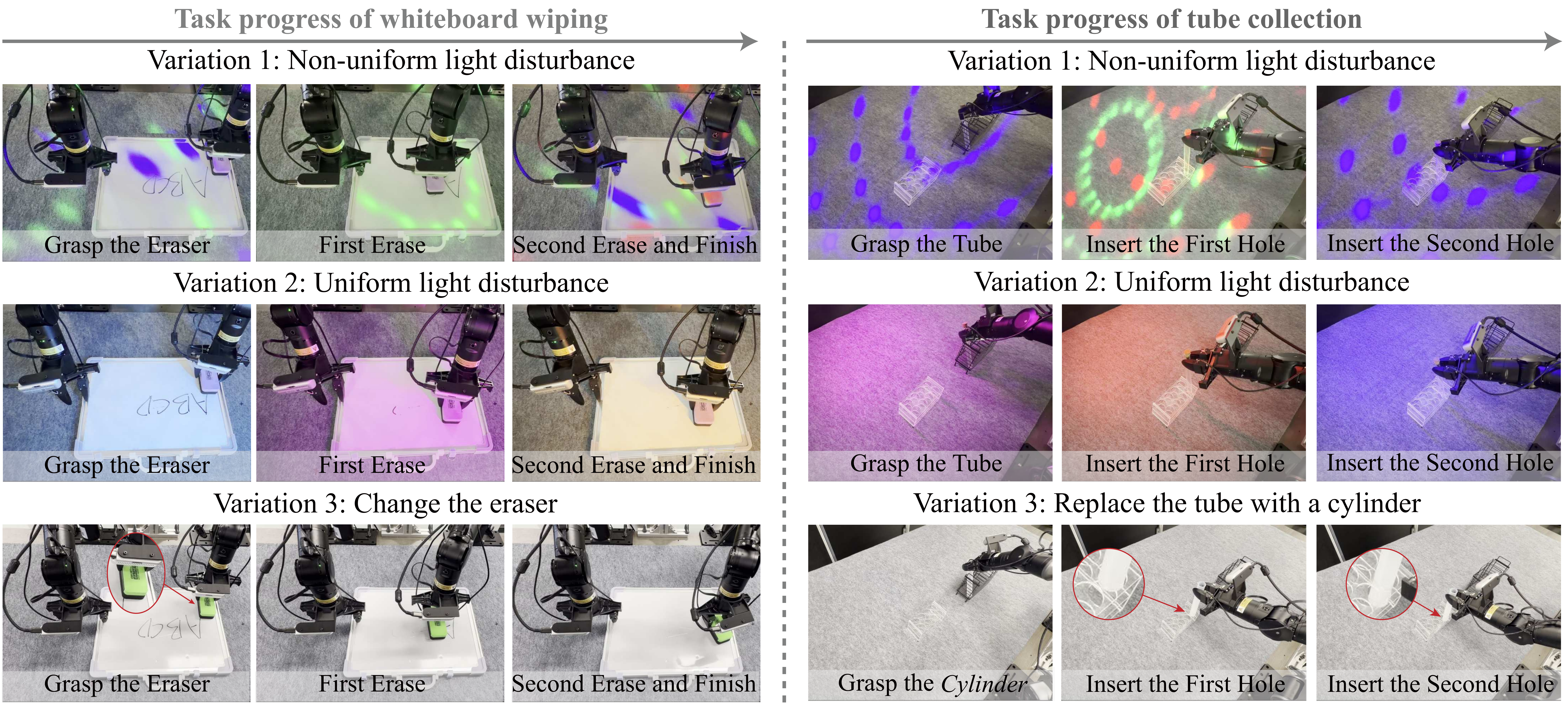}
  \caption{Robustness to environment and object variations. We evaluate the robustness of our method under both environment and object variations using  the whiteboard erasing and tube collection tasks. For environment variations, we introduce  temporal varying spatially non-uniform and uniform lighting conditions. For object variations, we change the eraser and ink appearance in the whiteboard erasing task, and replace the tube with a cylinder in the tube collection task. }
  \label{roubustness}
        \vspace{-2mm}

\end{figure*}

\noindent\textbf{How does TMC Facilitate Contact-rich Manipulation?}
We further validate the importance of TMC in contact-rich manipulation tasks. As shown in Table \ref{albation}, removing TMC leads to a performance drop in  whiteboard erasing and tube collection. For whiteboard erasing, the absence of TMC hinders the model's ability to implicitly infer force magnitudes, leading to inconsistent erasing pressure. For tube collection, without TMC, the model struggles to perceive fine-grained contact states leading to misalignment during insertion. These results demonstrate the effectiveness of TMC in facilitating contact-rich manipulation.

\noindent\textbf{How does MoT-based Strategy Improve Visuo-Tactile Manipulation?} Table \ref{albation} shows that replacing our MoT-based strategy with a simple concatenation operation leads to a performance drop, which validates the effectiveness of our MoT-based fusion framework in effectively integrating global visual perception and local tactile sensing for visuo-tactile manipulation tasks.

\noindent\textbf{Universality of TMC across Different Optical Sensors.} We further discuss the cross-sensor universality of TMC. Our experiments use two kinds of tactile sensors with different marker patterns \cite{XenseSensor,DM-TacW2,InTacS1}. Table \ref{tab:comparison} shows that our method achieves superior performance with various optical tactile  sensors, validating the universality of TMC. This advantage primarily stems from our motion-aware tactile representation, which prioritizes underlying gel deformation over raw visual appearance.

\noindent\textbf{Can TMC be Flexibly Embedded into Existing Policies?} To further validate the flexibility and effectiveness of TMC, we embed existing  representations (raw appearance and cumulative motion) and TMC into existing visual policies: ACT \cite{zhao2023learning} and DP \cite{chi2025diffusion} on the whiteboard erasing  task.  Table \ref{embedded} shows that existing methods obtain better performance after integrating the TMC, further  demonstrating the effectiveness of TMC.

\noindent\textbf{Effectiveness of MoT-based Visuo-Tactile Fusion Strategy.} We further study the effectiveness of our MoT-based fusion strategy by replacing it with several commonly used fusion strategies, including simple concatenation, cross attention. As shown in Table \ref{Text}, our MoT-based fusion strategy outperforms these commonly used fusion strategies, which validates the effectiveness of our MoT-based fusion framework in effectively integrating global visual perception and local tactile sensing for visuo-tactile manipulation tasks.

\noindent\textbf{Robustness to Environment and Object Variations.}
To evaluate the robustness of our method, we introduce various disturbances to the environment, including different temporal-varying lighting disturbances (e.g., spatial uniform and non-uniform) and object variations (e.g., different colored eraser and replace the tube with a cylinder). Figure \ref{roubustness} shows that our method still successfully completes the tasks under these disturbances, demonstrating its robustness to environmental and object variations.

\noindent\textbf{Limitations.} 
Though our method achieves fine-grained contact state perception and effective visuo-tactile fusion for contact-rich manipulation, its generalization remains limited under extreme visual disturbances or object position variations. Failure cases under such conditions are provided in the Appendix. In future work, we will focus on improving the generalization of contact-rich manipulation policies in complex environments.

\section{Conclusion}
We reveal that the correlation between transient and cumulative tactile motion provides explicit discrimination of fine-grained contact states. This insight motivates us to propose a motion-aware tactile representation to facilitate contact-rich manipulation. We introduce a unified modality-specific visuo-tactile policy that captures cross-modal complementarity while preserving the unique characteristics of each modality—thereby enabling the effective integration of global visual perception and local tactile feedback. Extensive experiments on a diverse range of tasks demonstrate the superior performance of our proposed method.

\section*{Acknowledgments}
This work is supported by the National Natural Science Foundation of China (Grant No. 62472098),  the Science and Technology Commission of Shanghai Municipality (No. 24511103100) and the New Cornerstone Science Foundation through the XPLORER PRIZE.

\clearpage

\bibliographystyle{plainnat}
\bibliography{main_v3}

@String(ICRA ={IEEE international conference on robotics and automation})

@String(ICRA ={ICRA})

@article{zhao2023learning,
  title={Learning fine-grained bimanual manipulation with low-cost hardware},
  author={Zhao, Tony Z and Kumar, Vikash and Levine, Sergey and Finn, Chelsea},
  journal={arXiv preprint arXiv:2304.13705},
  year={2023}
}

@article{zhao2024aloha,
  title={Aloha unleashed: A simple recipe for robot dexterity},
  author={Zhao, Tony Z and Tompson, Jonathan and Driess, Danny and Florence, Pete and Ghasemipour, Kamyar and Finn, Chelsea and Wahid, Ayzaan},
  journal={arXiv preprint arXiv:2410.13126},
  year={2024}
}

@article{chi2025diffusion,
  title={Diffusion policy: Visuomotor policy learning via action diffusion},
  author={Chi, Cheng and Xu, Zhenjia and Feng, Siyuan and Cousineau, Eric and Du, Yilun and Burchfiel, Benjamin and Tedrake, Russ and Song, Shuran},
  journal={The International Journal of Robotics Research},
  volume={44},
  number={10-11},
  pages={1684--1704},
  year={2025},
  publisher={Sage Publications Sage UK: London, England}
}

@article{gu2025tactilealoha,
  title={TactileAloha: Learning Bimanual Manipulation with Tactile Sensing},
  author={Gu, Ningquan and Kosuge, Kazuhiro and Hayashibe, Mitsuhiro},
  journal={IEEE Robotics and Automation Letters},
  year={2025},
  publisher={IEEE}
}

@article{van2024built,
  title={Built different: Tactile perception to overcome cross-embodiment capability differences in collaborative manipulation},
  author={van den Bogert, William and Iyengar, Madhavan and Fazeli, Nima},
  journal={arXiv e-prints},
  pages={arXiv--2409},
  year={2024}
}

@article{xue2025reactive,
  title={Reactive diffusion policy: Slow-fast visual-tactile policy learning for contact-rich manipulation},
  author={Xue, Han and Ren, Jieji and Chen, Wendi and Zhang, Gu and Fang, Yuan and Gu, Guoying and Xu, Huazhe and Lu, Cewu},
  journal={arXiv preprint arXiv:2503.02881},
  year={2025}
}

@article{wu2025freetacman,
  title={Freetacman: Robot-free visuo-tactile data collection system for contact-rich manipulation},
  author={Wu, Longyan and Yu, Checheng and Ren, Jieji and Chen, Li and Jiang, Yufei and Huang, Ran and Gu, Guoying and Li, Hongyang},
  journal={arXiv preprint arXiv:2506.01941},
  year={2025}
}

@article{yu2023mimictouch,
  title={Mimictouch: Leveraging multi-modal human tactile demonstrations for contact-rich manipulation},
  author={Yu, Kelin and Han, Yunhai and Wang, Qixian and Saxena, Vaibhav and Xu, Danfei and Zhao, Ye},
  journal={arXiv preprint arXiv:2310.16917},
  year={2023}
}

@article{li2025simultaneous,
  title={Simultaneous Tactile-Visual Perception for Learning Multimodal Robot Manipulation},
  author={Li, Yuyang and Chen, Yinghan and Zhao, Zihang and Li, Puhao and Liu, Tengyu and Huang, Siyuan and Zhu, Yixin},
  journal={arXiv preprint arXiv:2512.09851},
  year={2025}
}

@article{xu2025exumi,
  title={exUMI: Extensible Robot Teaching System with Action-aware Task-agnostic Tactile Representation},
  author={Xu, Yue and Wei, Litao and An, Pengyu and Zhang, Qingyu and Li, Yong-Lu},
  journal={arXiv preprint arXiv:2509.14688},
  year={2025}
}

@article{cheng2025omnivtla,
  title={Omnivtla: Vision-tactile-language-action model with semantic-aligned tactile sensing},
  author={Cheng, Zhengxue and Zhang, Yiqian and Zhang, Wenkang and Li, Haoyu and Wang, Keyu and Song, Li and Zhang, Hengdi},
  journal={arXiv preprint arXiv:2508.08706},
  year={2025}
}

@article{zhang2025vtla,
  title={Vtla: Vision-tactile-language-action model with preference learning for insertion manipulation},
  author={Zhang, Chaofan and Hao, Peng and Cao, Xiaoge and Hao, Xiaoshuai and Cui, Shaowei and Wang, Shuo},
  journal={arXiv preprint arXiv:2505.09577},
  year={2025}
}

@article{bi2025vla,
  title={Vla-touch: Enhancing vision-language-action models with dual-level tactile feedback},
  author={Bi, Jianxin and Ma, Kevin Yuchen and Hao, Ce and Shou, Mike Zheng and Soh, Harold},
  journal={arXiv preprint arXiv:2507.17294},
  year={2025}
}

@article{liu2025vitamin,
  title={Vitamin: Learning contact-rich tasks through robot-free visuo-tactile manipulation interface},
  author={Liu, Fangchen and Li, Chuanyu and Qin, Yihua and Xu, Jing and Abbeel, Pieter and Chen, Rui},
  journal={arXiv preprint arXiv:2504.06156},
  year={2025}
}

@article{zhu2025residual,
  title={Residual Rotation Correction using Tactile Equivariance},
  author={Zhu, Yizhe and Ye, Zhang and Hu, Boce and Zhao, Haibo and Qi, Yu and Wang, Dian and Platt, Robert},
  journal={arXiv preprint arXiv:2511.07381},
  year={2025}
}

@inproceedings{george2025vital,
  title={Vital pretraining: Visuo-tactile pretraining for tactile and non-tactile manipulation policies},
  author={George, Abraham and Gano, Selam and Katragadda, Pranav and Farimani, Amir Barati},
  booktitle={2025 IEEE International Conference on Robotics and Automation (ICRA)},
  pages={258--264},
  year={2025},
  organization={IEEE}
}

@article{zhang2026touchguide,
  title={TouchGuide: Inference-Time Steering of Visuomotor Policies via Touch Guidance},
  author={Zhang, Zhemeng and Ma, Jiahua and Yang, Xincheng and Wen, Xin and Zhang, Yuzhi and Li, Boyan and Qin, Yiran and Liu, Jin and Zhao, Can and Kang, Li and others},
  journal={arXiv preprint arXiv:2601.20239},
  year={2026}
}

@article{li2022see,
  title={See, hear, and feel: Smart sensory fusion for robotic manipulation},
  author={Li, Hao and Zhang, Yizhi and Zhu, Junzhe and Wang, Shaoxiong and Lee, Michelle A and Xu, Huazhe and Adelson, Edward and Fei-Fei, Li and Gao, Ruohan and Wu, Jiajun},
  journal={arXiv preprint arXiv:2212.03858},
  year={2022}
}

@article{huang2026tactile,
  title={Tactile-Force Alignment in Vision-Language-Action Models for Force-aware Manipulation},
  author={Huang, Yuzhe and Lin, Pei and Li, Wanlin and Li, Daohan and Li, Jiajun and Jiang, Jiaming and Xiao, Chenxi and Jiao, Ziyuan},
  journal={arXiv preprint arXiv:2601.20321},
  year={2026}
}

@article{huang2025vt,
  title={Vt-refine: Learning bimanual assembly with visuo-tactile feedback via simulation fine-tuning},
  author={Huang, Binghao and Xu, Jie and Akinola, Iretiayo and Yang, Wei and Sundaralingam, Balakumar and O'Flaherty, Rowland and Fox, Dieter and Wang, Xiaolong and Mousavian, Arsalan and Chao, Yu-Wei and others},
  journal={arXiv preprint arXiv:2510.14930},
  year={2025}
}

@article{li2025adaptive,
  title={Adaptive Visuo-Tactile Fusion with Predictive Force Attention for Dexterous Manipulation},
  author={Li, Jinzhou and Wu, Tianhao and Zhang, Jiyao and Chen, Zeyuan and Jin, Haotian and Wu, Mingdong and Shen, Yujun and Yang, Yaodong and Dong, Hao},
  journal={arXiv preprint arXiv:2505.13982},
  year={2025}
}

@article{zhu2025touch,
  title={Touch in the wild: Learning fine-grained manipulation with a portable visuo-tactile gripper},
  author={Zhu, Xinyue and Huang, Binghao and Li, Yunzhu},
  journal={arXiv preprint arXiv:2507.15062},
  year={2025}
}

@article{chen2025multi,
  title={Multi-Modal Manipulation via Multi-Modal Policy Consensus},
  author={Chen, Haonan and Xu, Jiaming and Chen, Hongyu and Hong, Kaiwen and Huang, Binghao and Liu, Chaoqi and Mao, Jiayuan and Li, Yunzhu and Du, Yilun and Driggs-Campbell, Katherine},
  journal={arXiv preprint arXiv:2509.23468},
  year={2025}
}

@article{huang20243d,
  title={3d-vitac: Learning fine-grained manipulation with visuo-tactile sensing},
  author={Huang, Binghao and Wang, Yixuan and Yang, Xinyi and Luo, Yiyue and Li, Yunzhu},
  journal={arXiv preprint arXiv:2410.24091},
  year={2024}
}

@article{liu2025mla,
  title={Mla: A multisensory language-action model for multimodal understanding and forecasting in robotic manipulation},
  author={Liu, Zhuoyang and Liu, Jiaming and Xu, Jiadong and Han, Nuowei and Gu, Chenyang and Chen, Hao and Zhou, Kaichen and Zhang, Renrui and Hsieh, Kai Chin and Wu, Kun and others},
  journal={arXiv preprint arXiv:2509.26642},
  year={2025}
}

@article{chen2025implicitrdp,
  title={ImplicitRDP: An End-to-End Visual-Force Diffusion Policy with Structural Slow-Fast Learning},
  author={Chen, Wendi and Xue, Han and Wang, Yi and Zhou, Fangyuan and Lv, Jun and Jin, Yang and Tang, Shirun and Wen, Chuan and Lu, Cewu},
  journal={arXiv preprint arXiv:2512.10946},
  year={2025}
}

@article{lee2025manipforce,
  title={ManipForce: Force-Guided Policy Learning with Frequency-Aware Representation for Contact-Rich Manipulation},
  author={Lee, Geonhyup and Lee, Yeongjin and Kim, Kangmin and Lee, Seongju and Noh, Sangjun and Back, Seunghyeok and Lee, Kyoobin},
  journal={arXiv preprint arXiv:2509.19047},
  year={2025}
}

@article{zhao2025touch,
  title={Touch begins where vision ends: Generalizable policies for contact-rich manipulation},
  author={Zhao, Zifan and Haldar, Siddhant and Cui, Jinda and Pinto, Lerrel and Bhirangi, Raunaq},
  journal={arXiv preprint arXiv:2506.13762},
  year={2025}
}

@article{yu2025forcevla,
  title={ForceVLA: Enhancing VLA Models with a Force-aware MoE for Contact-rich Manipulation},
  author={Yu, Jiawen and Liu, Hairuo and Yu, Qiaojun and Ren, Jieji and Hao, Ce and Ding, Haitong and Huang, Guangyu and Huang, Guofan and Song, Yan and Cai, Panpan and others},
  journal={arXiv preprint arXiv:2505.22159},
  year={2025}
}

@article{huang2025tactile,
  title={Tactile-VLA: unlocking vision-language-action model's physical knowledge for tactile generalization},
  author={Huang, Jialei and Wang, Shuo and Lin, Fanqi and Hu, Yihang and Wen, Chuan and Gao, Yang},
  journal={arXiv preprint arXiv:2507.09160},
  year={2025}
}

@article{liu2025factr,
  title={Factr: Force-attending curriculum training for contact-rich policy learning},
  author={Liu, Jason Jingzhou and Li, Yulong and Shaw, Kenneth and Tao, Tony and Salakhutdinov, Ruslan and Pathak, Deepak},
  journal={arXiv preprint arXiv:2502.17432},
  year={2025}
}

@inproceedings{wu2025canonical,
  title={Canonical representation and force-based pretraining of 3d tactile for dexterous visuo-tactile policy learning},
  author={Wu, Tianhao and Li, Jinzhou and Zhang, Jiyao and Wu, Mingdong and Dong, Hao},
  booktitle={2025 IEEE International Conference on Robotics and Automation (ICRA)},
  pages={6786--6792},
  year={2025},
  organization={IEEE}
}

@article{he2025foar,
  title={FoAR: Force-Aware Reactive Policy for Contact-Rich Robotic Manipulation},
  author={He, Zihao and Fang, Hongjie and Chen, Jingjing and Fang, Hao-Shu and Lu, Cewu},
  journal={IEEE Robotics and Automation Letters},
  year={2025},
  publisher={IEEE}
}

@article{chen2022visuo,
  title={Visuo-tactile transformers for manipulation},
  author={Chen, Yizhou and Sipos, Andrea and Van der Merwe, Mark and Fazeli, Nima},
  journal={arXiv preprint arXiv:2210.00121},
  year={2022}
}

@article{yuan2017gelsight,
  title={Gelsight: High-resolution robot tactile sensors for estimating geometry and force},
  author={Yuan, Wenzhen and Dong, Siyuan and Adelson, Edward H},
  journal={Sensors},
  volume={17},
  number={12},
  pages={2762},
  year={2017},
  publisher={MDPI}
}

@article{lin20239dtact,
  title={9dtact: A compact vision-based tactile sensor for accurate 3d shape reconstruction and generalizable 6d force estimation},
  author={Lin, Changyi and Zhang, Han and Xu, Jikai and Wu, Lei and Xu, Huazhe},
  journal={IEEE Robotics and Automation Letters},
  volume={9},
  number={2},
  pages={923--930},
  year={2023},
  publisher={IEEE}
}

@inproceedings{taylor2022gelslim,
  title={Gelslim 3.0: High-resolution measurement of shape, force and slip in a compact tactile-sensing finger},
  author={Taylor, Ian H and Dong, Siyuan and Rodriguez, Alberto},
  booktitle={2022 International Conference on Robotics and Automation (ICRA)},
  pages={10781--10787},
  year={2022},
  organization={IEEE}
}

@article{lin2022dtact,
  title={Dtact: A vision-based tactile sensor that measures high-resolution 3d geometry directly from darkness},
  author={Lin, Changyi and Lin, Ziqi and Wang, Shaoxiong and Xu, Huazhe},
  journal={arXiv preprint arXiv:2209.13916},
  year={2022}
}

@inproceedings{ren2023mc,
  title={MC-TAC: Modular camera-based tactile sensor for robot gripper},
  author={Ren, Jieji and Zou, Jiang and Gu, Guoying},
  booktitle={International Conference on Intelligent Robotics and Applications},
  pages={169--179},
  year={2023},
  organization={Springer}
}

@misc{DM-TacW2,
    title = {Daimon optical tactile sensor,
             DM-Tac W2},
    note  = {\url{https://www.dmrobot.com/product/p1/dm-tacw2.html}, 
             DM-Tac W2},
}

@misc{InTacS1,
    title = {Neote AI optical tactile sensor,
            InTac S1},
    note  = {\url{https://www.neoteai.com/}, 
             InTac S1},
}

@misc{XenseSensor,
    title = {Xense optical tactile sensor,
             XenseSensor},
    note  = {\url{https://www.xenserobotics.com/product/367/detail/9}, 
             XenseSensor},
}

@article{zhang2022finger,
  title={Finger-inspired rigid-soft hybrid tactile sensor with superior sensitivity at high frequency},
  author={Zhang, Jinhui and Yao, Haimin and Mo, Jiaying and Chen, Songyue and Xie, Yu and Ma, Shenglin and Chen, Rui and Luo, Tao and Ling, Weisong and Qin, Lifeng and others},
  journal={Nature communications},
  volume={13},
  number={1},
  pages={5076},
  year={2022},
  publisher={Nature Publishing Group UK London}
}

@article{liu2022neuro,
  title={Neuro-inspired electronic skin for robots},
  author={Liu, Fengyuan and Deswal, Sweety and Christou, Adamos and Sandamirskaya, Yulia and Kaboli, Mohsen and Dahiya, Ravinder},
  journal={Science robotics},
  volume={7},
  number={67},
  pages={eabl7344},
  year={2022},
  publisher={American Association for the Advancement of Science}
}

@article{liu2022printed,
  title={Printed synaptic transistor--based electronic skin for robots to feel and learn},
  author={Liu, Fengyuan and Deswal, Sweety and Christou, Adamos and Shojaei Baghini, Mahdieh and Chirila, Radu and Shakthivel, Dhayalan and Chakraborty, Moupali and Dahiya, Ravinder},
  journal={Science Robotics},
  volume={7},
  number={67},
  pages={eabl7286},
  year={2022},
  publisher={American Association for the Advancement of Science}
}

@article{jamone2015highly,
  title={Highly sensitive soft tactile sensors for an anthropomorphic robotic hand},
  author={Jamone, Lorenzo and Natale, Lorenzo and Metta, Giorgio and Sandini, Giulio},
  journal={IEEE sensors Journal},
  volume={15},
  number={8},
  pages={4226--4233},
  year={2015},
  publisher={IEEE}
}

@article{bhirangi2021reskin,
  title={Reskin: versatile, replaceable, lasting tactile skins},
  author={Bhirangi, Raunaq and Hellebrekers, Tess and Majidi, Carmel and Gupta, Abhinav},
  journal={arXiv preprint arXiv:2111.00071},
  year={2021}
}

@article{luo2025tactile,
  title={Tactile robotics: An outlook},
  author={Luo, Shan and Lepora, Nathan F and Yuan, Wenzhen and Althoefer, Kaspar and Cheng, Gordon and Dahiya, Ravinder},
  journal={IEEE Transactions on Robotics},
  year={2025},
  publisher={IEEE}
}

@article{jiang2025rotipbot,
  title={RoTipBot: Robotic handling of thin and flexible objects using rotatable tactile sensors},
  author={Jiang, Jiaqi and Zhang, Xuyang and Gomes, Daniel Fernandes and Do, Thanh-Toan and Luo, Shan},
  journal={IEEE Transactions on Robotics},
  year={2025},
  publisher={IEEE}
}

@article{li2024vision,
  title={When vision meets touch: A contemporary review for visuotactile sensors from the signal processing perspective},
  author={Li, Shoujie and Wang, Zihan and Wu, Changsheng and Li, Xiang and Luo, Shan and Fang, Bin and Sun, Fuchun and Zhang, Xiao-Ping and Ding, Wenbo},
  journal={IEEE Journal of Selected Topics in Signal Processing},
  volume={18},
  number={3},
  pages={267--287},
  year={2024},
  publisher={IEEE}
}

@article{helmholtz1858integrale,
  title={{\"U}ber Integrale der hydrodynamischen Gleichungen, welche den Wirbelbewegungen entsprechen.},
  author={Helmholtz, H von},
  year={1858},
  publisher={Walter de Gruyter, Berlin/New York Berlin, New York}
}

@article{abdi2010principal,
  title={Principal component analysis},
  author={Abdi, Herv{\'e} and Williams, Lynne J},
  journal={Wiley interdisciplinary reviews: computational statistics},
  volume={2},
  number={4},
  pages={433--459},
  year={2010},
  publisher={Wiley Online Library}
}

@article{liang2024mixture,
  title={Mixture-of-transformers: A sparse and scalable architecture for multi-modal foundation models},
  author={Liang, Weixin and Yu, Lili and Luo, Liang and Iyer, Srinivasan and Dong, Ning and Zhou, Chunting and Ghosh, Gargi and Lewis, Mike and Yih, Wen-tau and Zettlemoyer, Luke and others},
  journal={arXiv preprint arXiv:2411.04996},
  year={2024}
}

@article{wang2025hbridge,
  title={HBridge: H-Shape Bridging of Heterogeneous Experts for Unified Multimodal Understanding and Generation},
  author={Wang, Xiang and Zhang, Zhifei and Zhang, He and Lin, Zhe and Zhou, Yuqian and Liu, Qing and Zhang, Shiwei and Li, Yijun and Liu, Shaoteng and Zheng, Haitian and others},
  journal={arXiv preprint arXiv:2511.20520},
  year={2025}
}

@article{chen2025sam,
  title={Sam 3d: 3dfy anything in images},
  author={Chen, Xingyu and Chu, Fu-Jen and Gleize, Pierre and Liang, Kevin J and Sax, Alexander and Tang, Hao and Wang, Weiyao and Guo, Michelle and Hardin, Thibaut and Li, Xiang and others},
  journal={arXiv preprint arXiv:2511.16624},
  year={2025}
}

@article{deng2025emerging,
  title={Emerging properties in unified multimodal pretraining},
  author={Deng, Chaorui and Zhu, Deyao and Li, Kunchang and Gou, Chenhui and Li, Feng and Wang, Zeyu and Zhong, Shu and Yu, Weihao and Nie, Xiaonan and Song, Ziang and others},
  journal={arXiv preprint arXiv:2505.14683},
  year={2025}
}

@article{jin2025srum,
  title={Srum: Fine-grained self-rewarding for unified multimodal models},
  author={Jin, Weiyang and Niu, Yuwei and Liao, Jiaqi and Duan, Chengqi and Li, Aoxue and Gao, Shenghua and Liu, Xihui},
  journal={arXiv preprint arXiv:2510.12784},
  year={2025}
}

@article{li2026causal,
  title={Causal World Modeling for Robot Control},
  author={Li, Lin and Zhang, Qihang and Luo, Yiming and Yang, Shuai and Wang, Ruilin and Han, Fei and Yu, Mingrui and Gao, Zelin and Xue, Nan and Zhu, Xing and others},
  journal={arXiv preprint arXiv:2601.21998},
  year={2026}
}

@article{bi2025motus,
  title={Motus: A Unified Latent Action World Model},
  author={Bi, Hongzhe and Tan, Hengkai and Xie, Shenghao and Wang, Zeyuan and Huang, Shuhe and Liu, Haitian and Zhao, Ruowen and Feng, Yao and Xiang, Chendong and Rong, Yinze and others},
  journal={arXiv preprint arXiv:2512.13030},
  year={2025}
}

@article{huang2025motvla,
  title={MoTVLA: A Vision-Language-Action Model with Unified Fast-Slow Reasoning},
  author={Huang, Wenhui and Chen, Changhe and Qi, Han and Lv, Chen and Du, Yilun and Yang, Heng},
  journal={arXiv preprint arXiv:2510.18337},
  year={2025}
}

@article{cai2026internvla,
  title={InternVLA-A1: Unifying Understanding, Generation and Action for Robotic Manipulation},
  author={Cai, Junhao and Cai, Zetao and Cao, Jiafei and Chen, Yilun and He, Zeyu and Jiang, Lei and Li, Hang and Li, Hengjie and Li, Yang and Liu, Yufei and others},
  journal={arXiv preprint arXiv:2601.02456},
  year={2026}
}

@article{luo2026being,
  title={Being-H0. 5: Scaling Human-Centric Robot Learning for Cross-Embodiment Generalization},
  author={Luo, Hao and Wang, Ye and Zhang, Wanpeng and Zheng, Sipeng and Xi, Ziheng and Xu, Chaoyi and Xu, Haiweng and Yuan, Haoqi and Zhang, Chi and Wang, Yiqing and others},
  journal={arXiv preprint arXiv:2601.12993},
  year={2026}
}

@article{gu2025manualvla,
  title={ManualVLA: A Unified VLA Model for Chain-of-Thought Manual Generation and Robotic Manipulation},
  author={Gu, Chenyang and Liu, Jiaming and Chen, Hao and Huang, Runzhong and Wuwu, Qingpo and Liu, Zhuoyang and Li, Xiaoqi and Li, Ying and Zhang, Renrui and Jia, Peng and others},
  journal={arXiv preprint arXiv:2512.02013},
  year={2025}
}

@article{wu2026pragmatic,
  title={A Pragmatic VLA Foundation Model},
  author={Wu, Wei and Lu, Fan and Wang, Yunnan and Yang, Shuai and Liu, Shi and Wang, Fangjing and Zhu, Qian and Sun, He and Wang, Yong and Ma, Shuailei and others},
  journal={arXiv preprint arXiv:2601.18692},
  year={2026}
}

@inproceedings{kroeger2016fast,
  title={Fast optical flow using dense inverse search},
  author={Kroeger, Till and Timofte, Radu and Dai, Dengxin and Van Gool, Luc},
  booktitle={European conference on computer vision},
  pages={471--488},
  year={2016},
  organization={Springer}
}

@article{kingma2014adam,
  title={Adam: A method for stochastic optimization},
  author={Kingma, Diederik P and Ba, Jimmy},
  journal={arXiv preprint arXiv:1412.6980},
  year={2014}
}

@article{kang2026learning,
  title={Learning Force-Regulated Manipulation with a Low-Cost Tactile-Force-Controlled Gripper},
  author={Kang, Xuhui and Tian, Tongxuan and Lee, Sung-Wook and Huang, Binghao and Li, Yunzhu and Kuo, Yen-Ling},
  journal={arXiv preprint arXiv:2602.10013},
  year={2026}
}

@article{feng2026anytouch,
  title={Anytouch 2: General optical tactile representation learning for dynamic tactile perception},
  author={Feng, Ruoxuan and Zhou, Yuxuan and Mei, Siyu and Zhou, Dongzhan and Wang, Pengwei and Cui, Shaowei and Fang, Bin and Yao, Guocai and Hu, Di},
  journal={arXiv preprint arXiv:2602.09617},
  year={2026}
}

@article{feng2025anytouch,
  title={Anytouch: Learning unified static-dynamic representation across multiple visuo-tactile sensors},
  author={Feng, Ruoxuan and Hu, Jiangyu and Xia, Wenke and Gao, Tianci and Shen, Ao and Sun, Yuhao and Fang, Bin and Hu, Di},
  journal={arXiv preprint arXiv:2502.12191},
  year={2025}
}

@article{higuera2024sparsh,
  title={Sparsh: Self-supervised touch representations for vision-based tactile sensing},
  author={Higuera, Carolina and Sharma, Akash and Bodduluri, Chaithanya Krishna and Fan, Taosha and Lancaster, Patrick and Kalakrishnan, Mrinal and Kaess, Michael and Boots, Byron and Lambeta, Mike and Wu, Tingfan and others},
  journal={arXiv preprint arXiv:2410.24090},
  year={2024}
}

@article{feng2024play,
  title={Play to the score: Stage-guided dynamic multi-sensory fusion for robotic manipulation},
  author={Feng, Ruoxuan and Hu, Di and Ma, Wenke and Li, Xuelong},
  journal={arXiv preprint arXiv:2408.01366},
  year={2024}
}

@inproceedings{hogan2021seeing,
  title={Seeing through your skin: Recognizing objects with a novel visuotactile sensor},
  author={Hogan, Francois R and Jenkin, Michael and Rezaei-Shoshtari, Sahand and Girdhar, Yogesh and Meger, David and Dudek, Gregory},
  booktitle={Proceedings of the IEEE/CVF winter conference on applications of computer vision},
  pages={1218--1227},
  year={2021}
}

@article{ward2018tactip,
  title={The tactip family: Soft optical tactile sensors with 3d-printed biomimetic morphologies},
  author={Ward-Cherrier, Benjamin and Pestell, Nicholas and Cramphorn, Luke and Winstone, Benjamin and Giannaccini, Maria Elena and Rossiter, Jonathan and Lepora, Nathan F},
  journal={Soft robotics},
  volume={5},
  number={2},
  pages={216--227},
  year={2018},
  publisher={SAGE Publications Sage CA: Los Angeles, CA}
}

@inproceedings{yamaguchi2017implementing,
  title={Implementing tactile behaviors using fingervision},
  author={Yamaguchi, Akihiko and Atkeson, Christopher G},
  booktitle={2017 IEEE-RAS 17th International Conference on Humanoid Robotics (Humanoids)},
  pages={241--248},
  year={2017},
  organization={IEEE}
}

\clearpage
\beginappendix

\section{Overview}
\label{sec:appendix_intro}
This appendix is organized as follows:

(\textbf{Section \ref{sec:Method}}) We offer additional analysis and details about the proposed method.

\begin{itemize}[label=\textbullet]
  \item We provide more analysis of tactile motion correlation properties using various optical tactile sensors in Sec. \ref{tactile_motion_correlation}.
  \item We provide more details about the efficient optical flow algorithm for tactile motion estimation in Sec. \ref{optical_flow_algorithm}.
\end{itemize}

(\textbf{Section \ref{sec:Exp}}) We provide additional details about the experimental setup.

\begin{itemize}[label=\textbullet]
  \item We offer more training details about our model in Sec. \ref{Training_Details}.
  \item We provide more details about the robotic setup in Sec. \ref{Hardware_Details}.
\end{itemize}

(\textbf{Section \ref{sec:DIS_ANA}}) We conduct further discussions on the failure cases of baseline and the limitations of the proposed method.

\begin{itemize}[label=\textbullet]
  \item We provide more experimental visualization results with another dense-marker sensor in Sec. \ref{additional_experiments}.
  \item We discuss the failure cases of the baseline methods in Sec. \ref{failure}.
  \item We provide more details about the limitations of the proposed method in Sec. \ref{limitation}.
\end{itemize}

(\textbf{Demo}) We provide a video demo to show the superiority and robustness of the proposed method towards challenging contact-rich manipulation tasks.

\section{Method Details}
\label{sec:Method}

\subsection{Tactile Motion Correlation of Various Tactile Optical Sensors}
\label{tactile_motion_correlation}
\begin{figure*}[t]
  \centering
  \captionsetup{skip=1.3mm}
     \includegraphics[width=\linewidth]{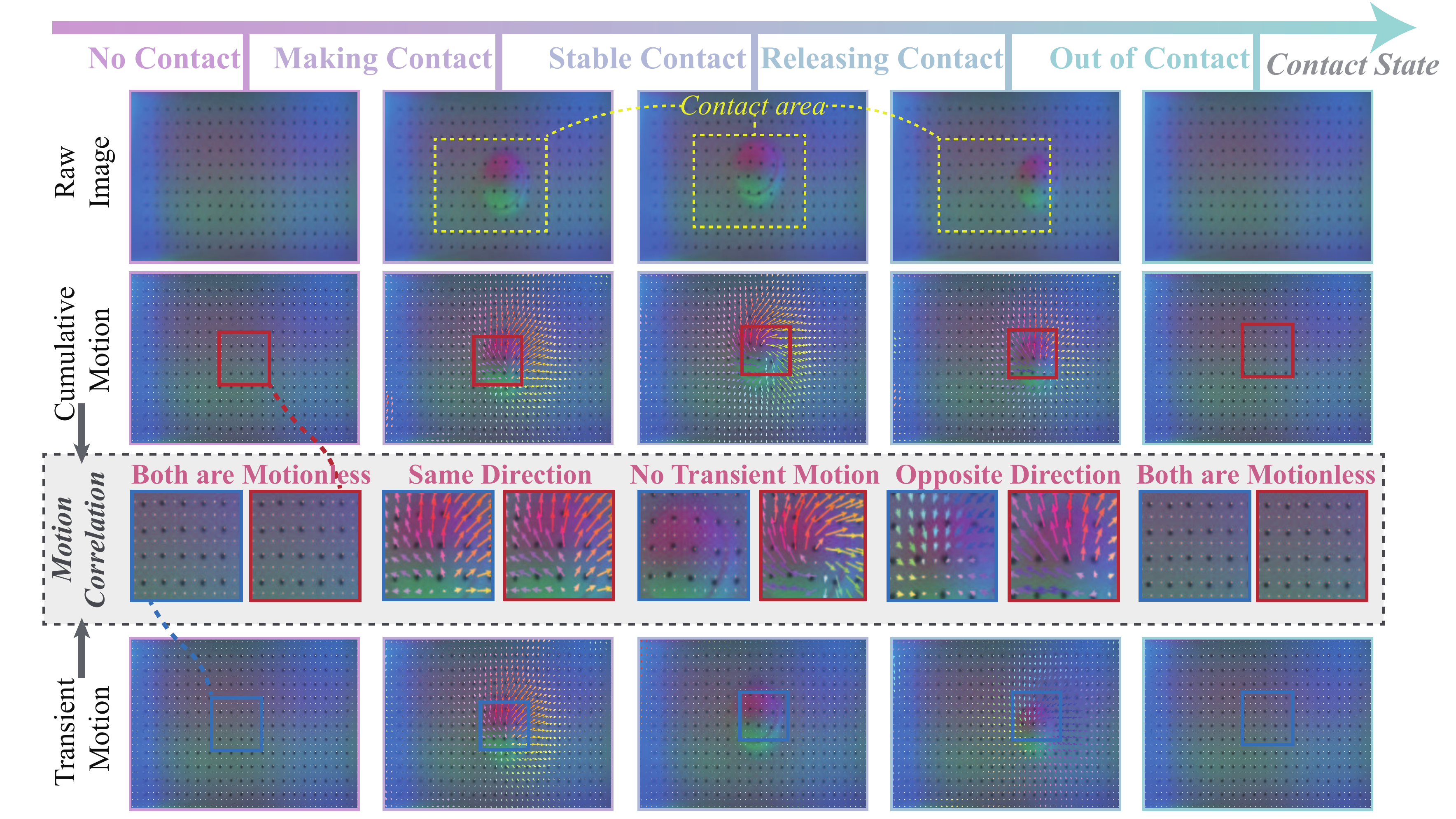}
  \caption{Analysis of tactile motion correlation properties using sparse-marker \cite{XenseSensor} optical tactile sensor.}
  \label{Tactile_motion_Xense}
\end{figure*}

\begin{figure*}[t]
  \centering
  \captionsetup{skip=1.3mm}
     \includegraphics[width=\linewidth]{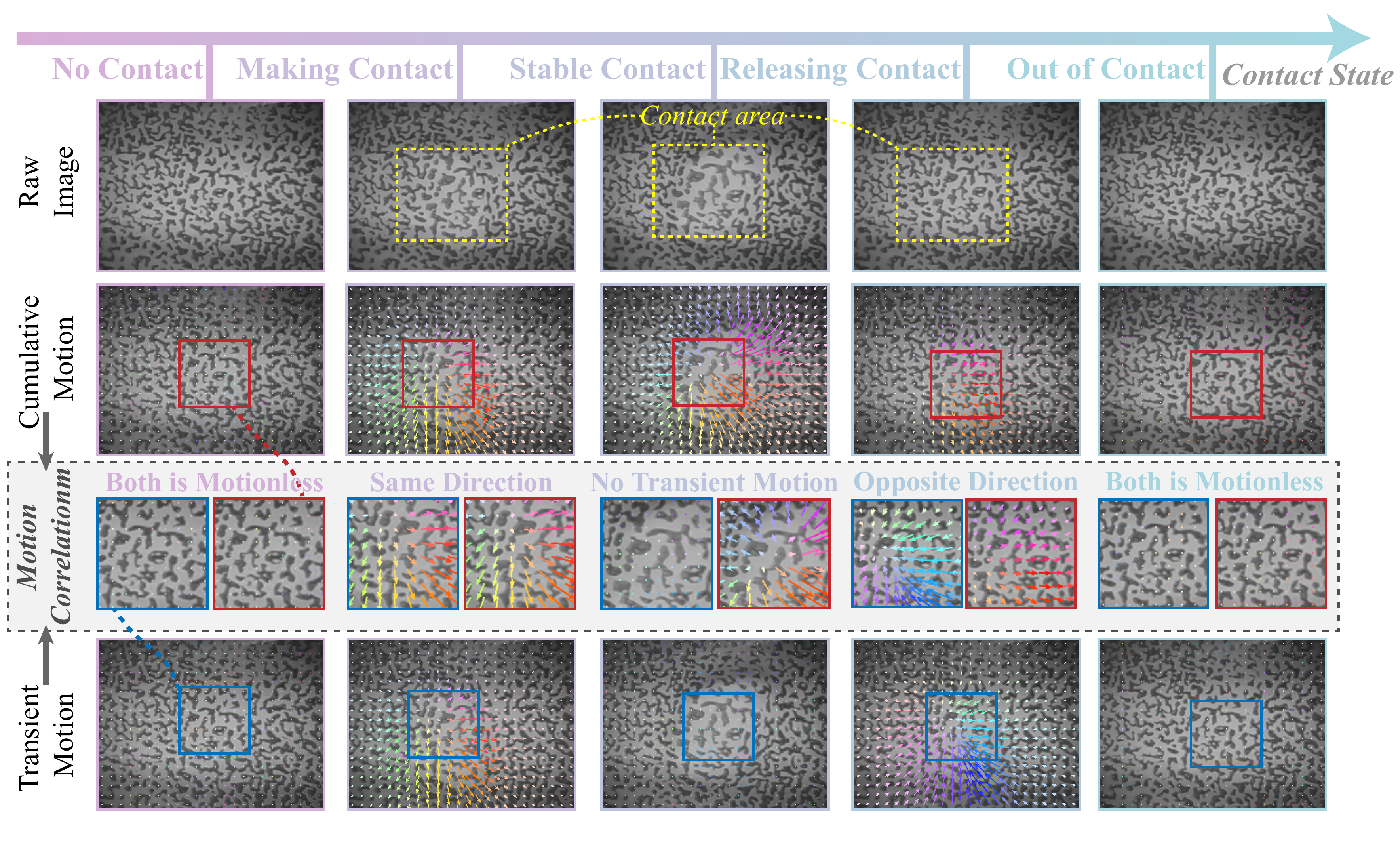}
  \caption{Analysis of tactile motion correlation properties using dense-marker \cite{DM-TacW2} optical tactile sensor.}
  \label{Tactile_motion_Lab}
\end{figure*}

To further validate the \textit{universality} of the proposed tactile motion correlation across different types of optical tactile sensors, we conduct analysis experiments on two kinds of optical sensors: one with regularly-arranged sparse markers \cite{XenseSensor} and one with random-distributed dense markers \cite{DM-TacW2}. We compute the transient motion and cumulative motion using efficient optical flow algorithm \cite{kroeger2016fast} for different contact states. As shown in Fig. \ref{Tactile_motion_Xense} and Fig. \ref{Tactile_motion_Lab}, we visualize the raw tactile images, transient motion, and cumulative motion for different contact states. We find that the motion correlations on both sensors exhibit distinctive discriminative power for identifying contact states, which further validates the universality of the proposed tactile motion correlation.

\subsection{Efficient Optical Flow Algorithm for Tactile Motion Estimation}
\label{optical_flow_algorithm}
To ensure a balance between computational efficiency and accuracy, we adopt the efficient optical flow algorithm DIS \cite{kroeger2016fast} based on inverse search to compute pixel motion between two tactile images. The algorithm can achieve a runtime of 600Hz on a single-core CPU at a resolution of 1024$\times$436, meeting the requirements for real-time manipulation and computation. The algorithm consists of three main stages: inverse search, multi-scale pyramid, and variational refinement. Table \ref{DIS} summarizes the parameter settings used in each stage.

\begin{table}[t]
\centering
\caption{DIS Optical Flow parameters categorized by algorithm stages.}
\begin{tabular}{ccc}
\hline
\textbf{Parameter} & \textbf{Description} & \textbf{Typical Value} \\
\hline
\multicolumn{3}{c}{\textbf{Patch Inverse Search}} \\
\hline
$\theta_{ps}$ & Patch size & 8 \\
$\theta_{ov}$ & Patch overlap ratio & 0.3 \\
$\theta_{it}$ & Inverse search iterations & 8 \\
\hline
\multicolumn{3}{c}{\textbf{Multi-scale Pyramid}} \\
\hline
$\theta_{sf}$ & Finest scale level & 2 \\
$\theta_{ss}$ & Coarsest scale level & 5 \\
$\theta_{sd}$ & Downscale factor & 2 \\
\hline
\multicolumn{3}{c}{\textbf{Variational Refinement}} \\
\hline
$\delta$ & Intensity consistency weight & 5 \\
$\gamma$ & Gradient consistency weight & 10 \\
$\alpha$ & Smoothness weight & 10 \\
$\theta_{vo}$ & Outer iterations & $s+1$ \\
$\theta_{vi}$ & Inner SOR iterations & 5 \\
\hline
\end{tabular}
\label{DIS}
\end{table}

\section{Experimental Details}
\label{sec:Exp}

\subsection{Training Details}
\label{Training_Details}

Our model is trained using a $\beta$-VAE objective to model the stochasticity in human demonstrations.
The overall loss function is defined as:
\begin{equation}
\mathcal{L} = \mathcal{L}_{\text{pred}} + \beta \mathcal{L}_{\text{KL}}.
\end{equation}
The prediction loss employs an L1 loss for precise action prediction:
\begin{equation}
\mathcal{L}_{\text{pred}} =
\left\| \hat{a}_{t:t+k} - a_{t:t+k} \right\|_1,
\end{equation}
where $\hat{a}_{t:t+k}$ denotes the predicted action chunk and $a_{t:t+k}$ denotes the ground-truth action sequence.
The KL loss regularizes the posterior distribution
\begin{equation}
q_\phi(z \mid a_{t:t+k}, o_t)
\end{equation}
toward a standard Gaussian prior:
\begin{equation}
\mathcal{L}_{\text{KL}} =
D_{\text{KL}}\left(
q_\phi(z \mid a_{t:t+k}, o_t)
\;\|\;
\mathcal{N}(0, I)
\right).
\end{equation}
The hyperparameter $\beta$ modulates the strength of the information bottleneck. The model is trained from scratch for each task using the Adam optimizer \cite{kingma2014adam}.
Training takes approximately 5 hours per task on a single RTX 3090 Ti GPU. The core training parameters are summarized in Table \ref{training_parameters}.

\begin{table*}[t]
\centering
\caption{Core training parameters of our model.}
\begin{tabular}{llll}
\hline
\textbf{Parameter} & \textbf{Value} & \textbf{Hyperparameter} & \textbf{Value} \\
\hline
Learning rate & $2 \times 10^{-4}$ & Batch size & 16 \\
\# encoder layers & 4 & \# decoder layers & 7 \\
Feedforward dimension & 3200 & Hidden dimension & 512 \\
\# heads & 8 & Chunk size ($k$) & 100 \\
Beta ($\beta$) & 10 & Dropout & 0.1 \\
\hline
\end{tabular}
\label{training_parameters}
\end{table*}

\subsection{Robotic Setup Details}
\label{Hardware_Details}
\begin{figure*}[b]
  \centering
  \captionsetup{skip=1.3mm}
     \includegraphics[width=\linewidth]{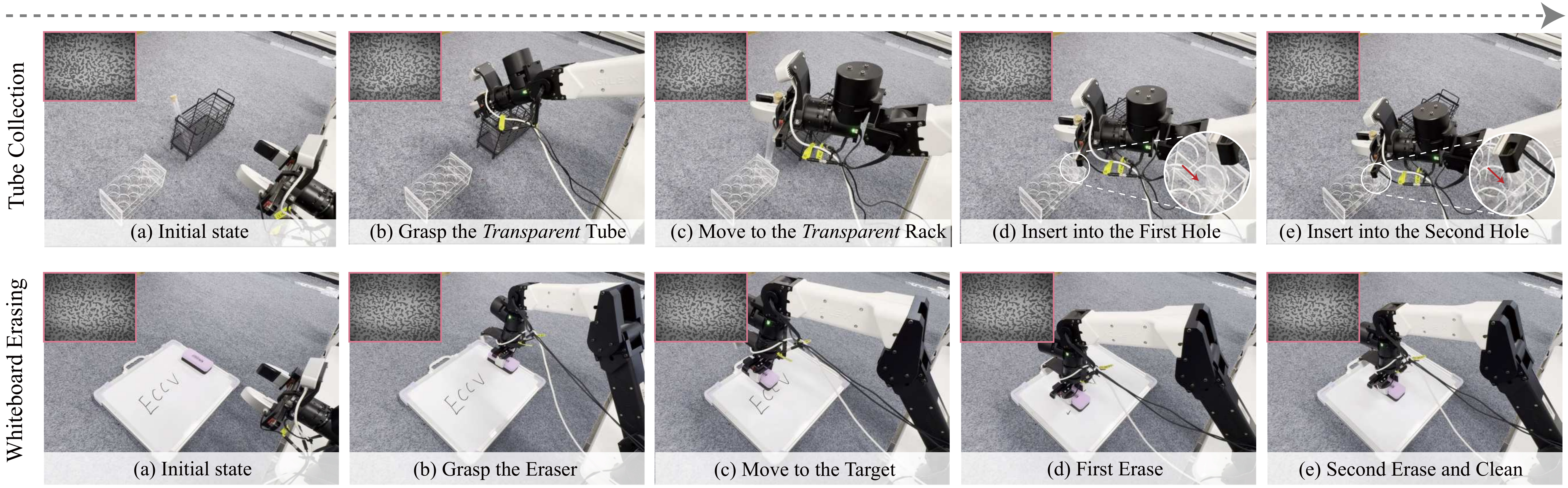}
  \caption{Real-world experiments with the another dense-marker optical tactile sensor \cite{DM-TacW2} on whiteboard wiping and tube collection.}
  \label{figure_daimon}
\end{figure*}

\begin{figure*}[t]
  \centering
  \captionsetup{skip=1.3mm}
     \includegraphics[width=\linewidth]{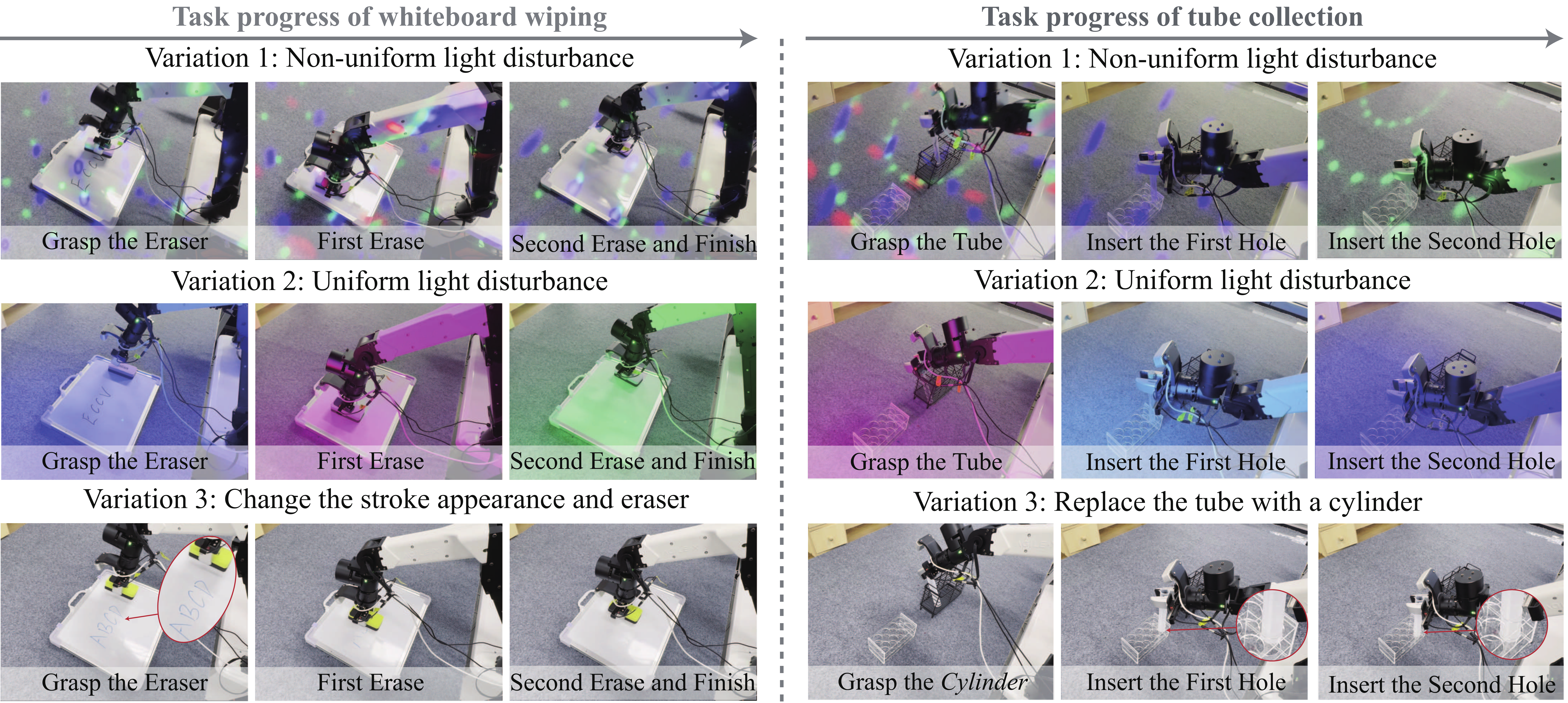}
  \caption{Robustness to environment and object variations with the another dense-marker optical tactile sensor \cite{DM-TacW2}. We evaluate the robustness of our method under both environment and object variations using the whiteboard erasing and tube collection tasks. For environment variations, we introduce temporally varying spatially non-uniform and uniform lighting conditions. For object variations, we change the eraser and ink appearance in the whiteboard erasing task, and replace the tube with a cylinder in the tube collection task.}
  \label{robustness}
\end{figure*}

\noindent\textbf{Robotic Setup.} The robotic system consists of two 6-DoF Agilex robotic arms each equipped with a 1-DoF parallel gripper. Tactile sensors are securely mounted on the end-effector using custom-designed adapters. The tactile sensors are connected to a high-speed data acquisition system that synchronizes tactile data with the robot's state information. The entire setup is controlled through a real-time operating system to ensure low-latency processing and actuation during manipulation tasks.

\noindent\textbf{Data Collection.} The data collection method is based on a master-slave teleoperation approach, where the master arm is operated by a human demonstrator to perform the desired manipulation tasks, while the slave arm replicates the movements of the master arm to collect synchronized tactile and state data for training the model.

\begin{figure*}[b]
  \centering
  \captionsetup{skip=1.3mm}
     \includegraphics[width=\linewidth]{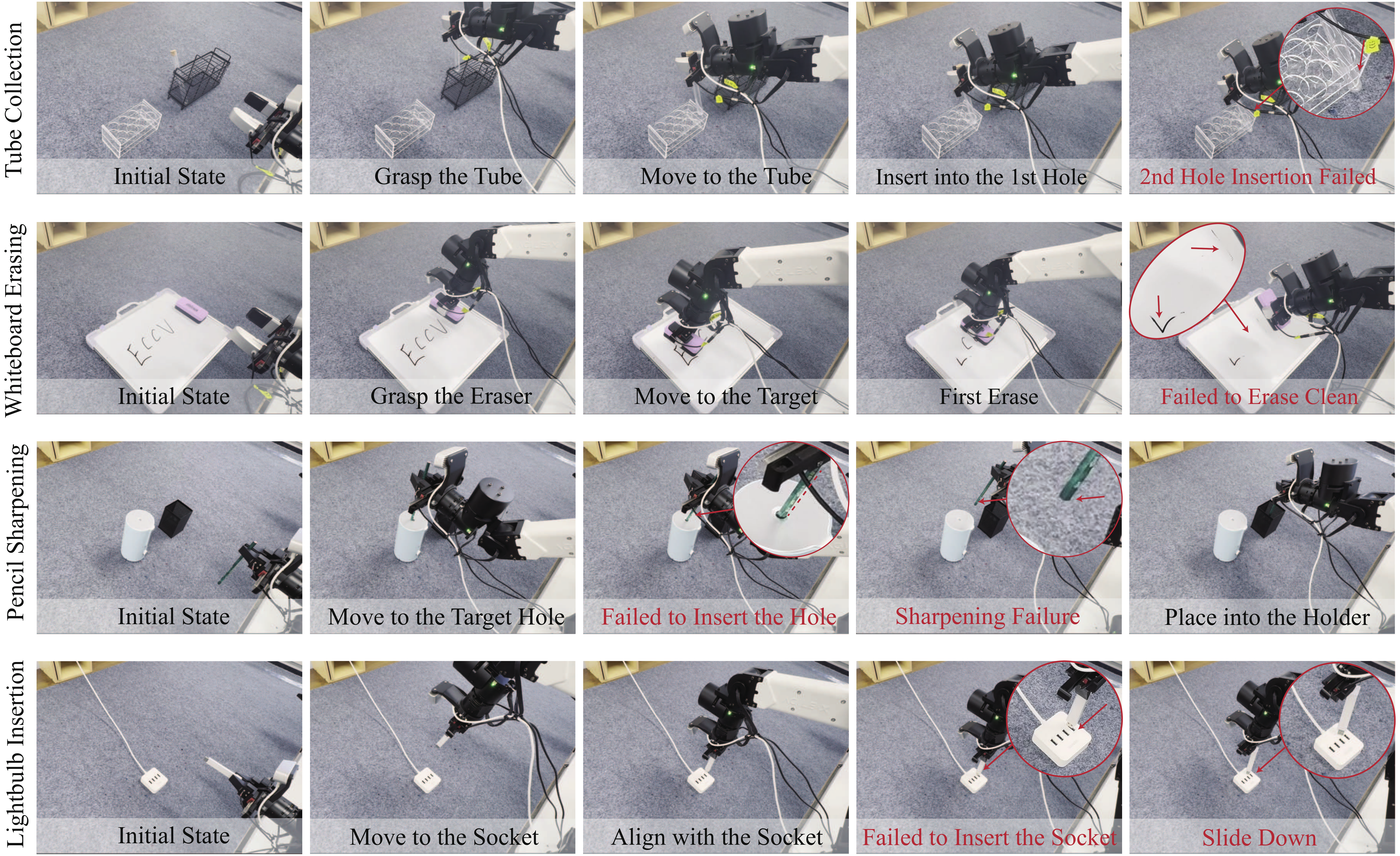}
  \caption{Visualization of baseline failure cases.}
  \label{failure_cases}
\end{figure*}

\begin{figure*}[t]
  \centering
  \captionsetup{skip=1.3mm}
     \includegraphics[width=\linewidth]{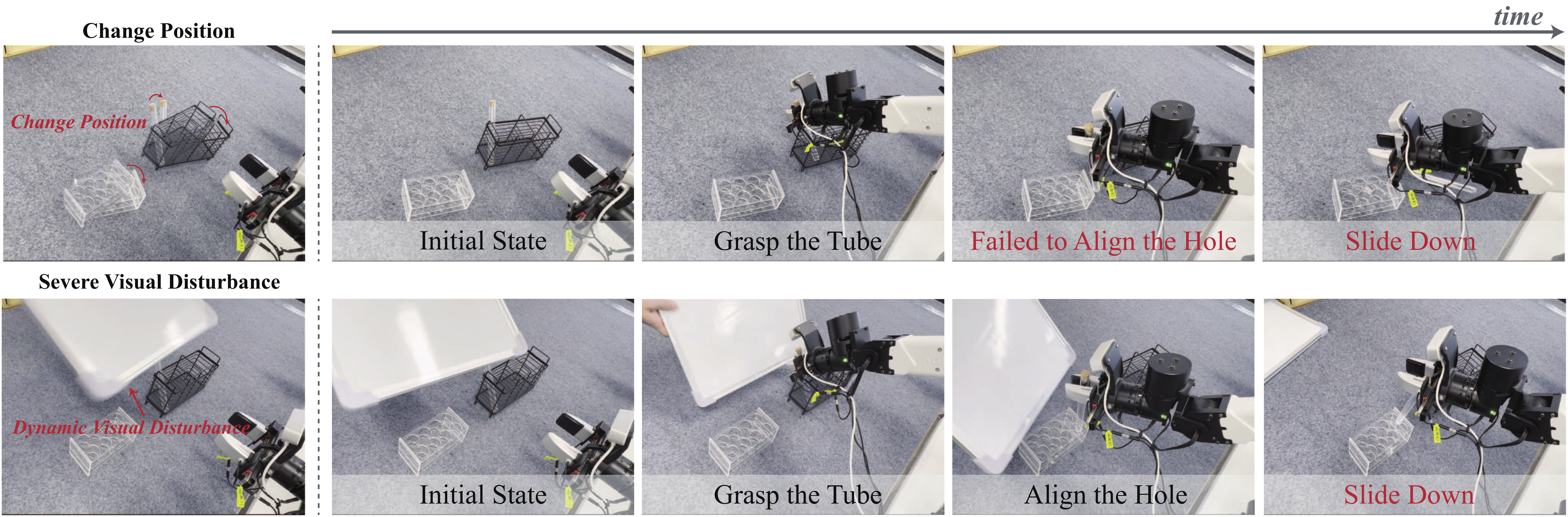}
  \caption{Limitations of the proposed method.}
  \label{Limitation}
\end{figure*}

\section{Discussion}
\label{sec:DIS_ANA}

\subsection{Additional Real-world Experiments}
\label{additional_experiments}
We further conduct real-world experiments on the another dense-marker optical tactile sensor for whiteboard wiping and tube collection tasks, as shown in Fig. \ref{figure_daimon}. We also evaluate the robustness of our method under varying lighting conditions and object replacements in Fig. \ref{robustness}. In the whiteboard wiping task, we introduce temporally varying spatially non-uniform and uniform lighting disturbances, while in the tube collection task, we replace the tube with a cylinder. The experimental results demonstrate that our method can successfully complete the tasks under these challenging conditions, validating its robustness.

\subsection{Baseline Failure Case Analysis}
\label{failure}
We further visualize the failure cases of the baseline method (visual-only) in Fig. \ref{failure_cases}. In the tube collection task, due to visual occlusion and lack of contact state perception, the baseline method fails to accurately perceive the position of the second hole, leading to insertion failure. In the whiteboard erasing task, the absence of tactile perception results in the baseline method's inability to sense the force magnitude, leading to incomplete erasing. In the lightbulb insertion task, due to the lack of contact state perception, the baseline method fails to detect insertion failures during the process, resulting in an inability to adjust actions in time and ultimately leading to insertion failure. In the pencil sharpening task, due to the lack of shear force and contact perception, the baseline method fails to apply appropriate force magnitude, resulting in inserting the pencil too shallowly and causing sharpening failure.

\subsection{Limitation and Future Work}
\label{limitation}
Although our method achieves superior fine-grained contact state perception and enables effective visuo-tactile fusion for contact-rich manipulation, its generalization remains limited under extreme visual disturbances or significant object pose variations. Failure cases under such conditions are shown in Fig. \ref{Limitation}.
We apply position changes and severe visual disturbances in the tube collection task. It can be observed that while our method can successfully grasp the tube under these conditions, it encounters failures during the insertion process.
In future work, we will focus on improving the generalization of contact-rich manipulation policies in complex environments.

\end{document}